\documentclass[lettersize,journal]{IEEEtran}
\usepackage{amsmath,amsfonts}
\usepackage{algorithmic}
\usepackage{algorithm}
\usepackage{array}
\usepackage{textcomp}
\usepackage{hyperref}       
\usepackage{stfloats}
\usepackage{url}
\usepackage{verbatim}
\usepackage{graphicx}
\usepackage{booktabs}       
\usepackage{amsfonts}       
\usepackage{nicefrac}       
\usepackage{microtype}      
\usepackage{xcolor}         
\usepackage{amsmath}
\usepackage{graphicx}
\usepackage{amsthm}
\usepackage{algorithmic}
\usepackage{algorithm}
\usepackage{multirow}
\usepackage{makecell}
\usepackage{pifont}
\usepackage{amssymb}
\begin{document}

\title{Causal Semantic Alignment for LLM-based Time Series Forecasting}


\author{Kexuan Zhang, Xiaobei Zou, Cesare Alippi, \IEEEmembership{Fellow, IEEE}, Gary G. Yen, \IEEEmembership{Fellow, IEEE}, and Yang Tang, \IEEEmembership{Fellow, IEEE}}



\maketitle
\begin{abstract}
Recent advances in Large Language Models (LLMs) have opened new possibilities for time series forecasting by enabling alignment between temporal patterns and pretrained word embeddings. However, most LLM-based methods overlook the heterogeneous nature of time series, where dynamic fluctuations and invariant semantics are entangled. This entanglement introduces spurious correlations during the alignment, as dynamic components act as confounders by simultaneously influencing invariant components and the resulting aligned embeddings. To address this issue, a variable-level alignment framework CVAformer (\textbf{\underline{C}}ausal \textbf{\underline{V}}ariable-level \textbf{\underline{A}}lignment Trans\textbf{\underline{former}}) is proposed. CVAformer explicitly disentangles each variable into invariant and dynamic components just before alignment, and applies causal intervention to mitigate the confounding effect of the dynamics. To better support variable-level alignment, CVAformer replaces the standard causal attention in LLMs with a non-causal attention mechanism that captures interactions among variables at each time step. Extensive experiments across long-term, short-term, few-shot, and zero-shot forecasting settings indicate that CVAformer matches or exceeds state-of-the-art performance on most datasets, and in some cases achieves notably better accuracy. Experimental results validate the effectiveness of variable-level alignment and dynamic disentanglement in CVAformer, offering a new perspective for LLM-based time series tasks.
\end{abstract}


\section{Introduction}
\IEEEPARstart{T}{ime} series forecasting plays a crucial role in a wide range of real-world applications, including climate science \cite{corrformer}, finance \cite{iTransformer}, energy systems \cite{10453229}, and traffic management \cite{9983531}. The increasing demand for general-purpose forecasting models that support variable-length inputs and integrate multimodal information has motivated recent efforts toward foundation models for time series with improved generalization abilities \cite{LLMTime,GPT4TS,11017752}.

The rapid progress of Large Language Models (LLMs) \cite{radford2019language,bert,touvron2023llama} has inspired the time series community too, owing to their ability to encode semantic structures and long-range dependencies. Pretrained on massive corpora, LLMs excel at contextual reasoning, making them suitable for cross-modal tasks across domains such as vision and biology \cite{visionllm,proteinllm}. Motivated by these features, early attempts \cite{GPT4TS,LLMTime} reformulate time series as linguistic sequences, enabling LLMs to perform end-to-end forecasting. Subsequent studies \cite{Autotimes,GPT4TS} emphasize the importance of modality alignment and propose explicit mapping strategies to better adapt time series to the language space. Meanwhile, \cite{TimeLLM, CALF} introduce pretrained word embeddings into LLM-based predictors to enhance semantic alignment and knowledge transfer.

\begin{figure*}[t]
    \centering
    \includegraphics[width=0.8\linewidth]{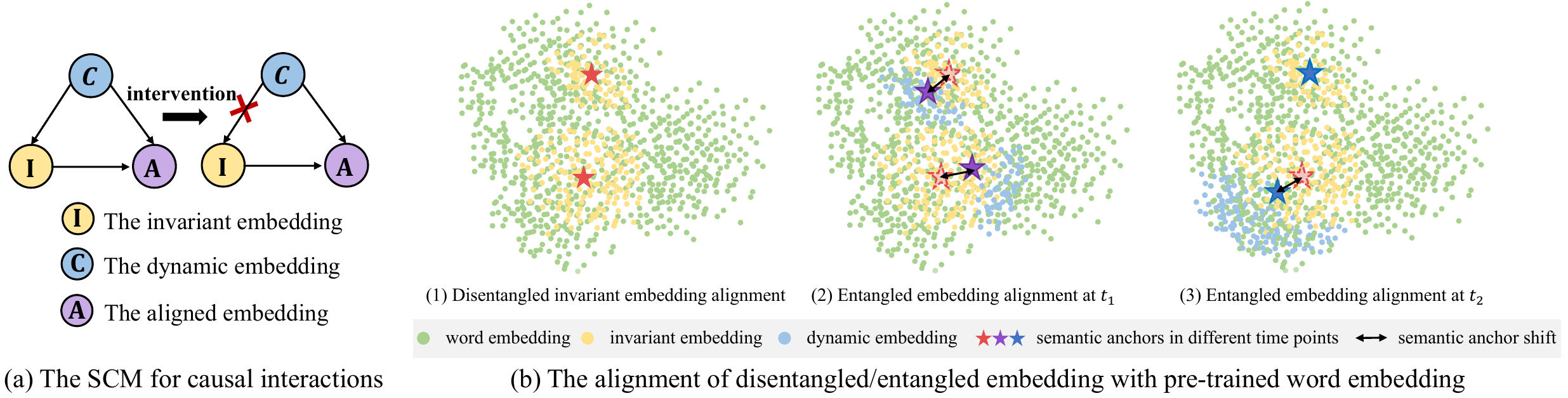}
    \caption{(a) The illustration of the alignment process between time embeddings and pretrained word embeddings in LLM-based time series forecasting. Due to ever-changing dynamic components, the semantic anchors of entangled embeddings may drift over time. (b) A structural causal model (SCM) depicting the causal relationships among the invariant component $I$, the dynamic component $C$, and the aligned embedding $A$. In this work, CVAformer applies backdoor intervention to eliminate the confounding effect of the dynamic on the invariant embedding and ensure unbiased alignment.}
    \label{motivation}
\end{figure*}

Existing LLM-based time-series processing methods, see, e.g., \cite{TimeLLM, CALF}, commonly construct a unified time embedding, and then align such embeddings with pretrained word embeddings. While this strategy yields promising results, it inherently entangles two fundamentally different factors: (1) invariant semantics that represent the intrinsic identity of each variable, and (2) dynamic components that reflect externally induced variations, which generally emerges as high-frequency fluctuations. From a causal perspective, the dynamic component acts as a confounder, as it simultaneously influences both the estimated invariant representation and the final aligned embedding; see Fig. \ref{motivation} (a). Although the true invariant semantics $I$ are expected to be constant over time, the observed invariant embedding is extracted from an entangled time embedding contaminated by dynamic variations $C$. As a result, fluctuations in $C$ generally shift the distribution of the entangled embedding even when the underlying semantics remain unchanged. This phenomenon leads to semantic anchor drifts, as shown in Fig. \ref{motivation} (b): the dynamic-induced shifts in the entangled embedding cause the alignment target to move over time, resulting in unstable and biased alignments. Consequently, the aligned embedding $A$ may fail to consistently capture the true semantic identity of each variable.

Furthermore, most existing LLM-based models for time series \cite{GPT4TS,TimeLLM,TEMPO} perform cross-modal alignment at the temporal level by aligning single time steps or preprocessed temporal patches with pretrained word embeddings. However, as pointed out in \cite{iTransformer}, treating each timestamp as a token yields unreliable semantics due to narrow receptive fields and temporal misalignment across variables. While patch-based approaches mitigate this issue by capturing broader temporal patterns at lower computational cost, the resulting embeddings are often less interpretable. Although \cite{CALF} introduces a channel-wise mapping to encode variables from the variable level, it overlooks the confounding effects introduced by dynamic components as discussed above. These limitations highlight the need for alignment strategies that operate at the variable level while accounting for the inevitable confounding during alignment. In addition, the causal attention mechanism inherent in LLMs for sequence modeling \cite{radford2019language} is not suitable for variable-level alignment, as variables do not obey any inherent ordering. Imposing autoregressive structure on unordered variables introduces artificial constraints and inductive biases that do not reflect true inter-variable relationships.

Motivated by the above observations, we propose CVAformer (\textbf{\underline{C}}ausal \textbf{\underline{V}}ariable-level \textbf{\underline{A}}lignment Trans\textbf{\underline{former}}) framework, which conducts the alignment from a variable-level with two coordinated branches: the temporal branch and the textual branch. Specifically, the temporal branch employs stacked Non-Causal Attention Blocks (NCBlocks) to capture inter-variable interactions without autoregressive constraints, while the textual branch conducts variable-level alignment with pretrained word embeddings under causal guidance. To address the entanglement between dynamic and invariant components, CVAformer decomposes each variable into a stable invariant part and a dynamic one, and performs a causal adjustment to isolate the effect of the invariant semantics. This is achieved by modeling the dynamic component as a confounder through a CausalEncoder and mitigating its influence via a soft gating mechanism. By combining variable-level non-causal temporal modeling and causal-aware alignment, CVAformer achieves more interpretable representations and improved generalization across diverse forecasting scenarios.

The novel contributions of this paper are summarized as:
\begin{itemize}
\item \textbf{A new variable-level alignment paradigm for LLM-based time series forecasting.} We introduce a variable-centric alignment framework that departs from traditional tokenizations and enables LLMs to capture the semantic information without imposing artificial temporal orderings.
\item \textbf{A causally grounded disentanglement methodology.} We provide a principled causal formulation that separates invariant semantics from dynamic fluctuations and introduces a backdoor-inspired mechanism to weaken confounding effects during cross-modal alignment.
\item \textbf{A unified architecture integrating non-causal attention for variable interactions.} We design a general modeling component based on non-causal attention that captures unordered inter-variable dependencies and can be readily incorporated into existing LLM-based predictors.
\item \textbf{Comprehensive empirical evidence demonstrating improved generalization.} Extensive evaluations across long/short-term, few/zero-shot forecasting settings show consistent improvements over state-of-the-art, highlighting the robustness and broad applicability of CVAformer.
\end{itemize}

The remainder of this paper is structured as follows: Section \ref{related work} introduces time series forecasting and cross-modal alignment for time series. Section \ref{method} details the proposed architecture and the overall workflow of CVAformer. Section \ref{experiment} introduces the experimental settings, the results of different forecasting tasks, and ablation studies. Finally, conclusions and future research directions are proposed in Section \ref{conclusion}.

\section{Related Work} \label{related work}
\subsection{Time Series Forecasting}
Time series forecasting is relevant in numerous applications, ranging from financial forecasting and weather prediction to activity recognition \cite{9768200,corrformer,iTransformer,11263981,10697287}. Traditional statistical models like ARIMA \cite{stevenson2007comparison}, VAR \cite{lutkepohl2013vector}, and MA \cite{6708545} offer principle-based approaches, but often struggle with high-dimensional, nonlinear, non-stationary data \cite{liu2022non}. These limitations have motivated the development of modern data-driven forecasting methods, including local and global deep models, graph-based architectures, transformer-based forecasters, and more recently, LLM-driven time-series predictors.

Recent advances in deep time-series forecasting have been driven by increasingly expressive model families. Transformer-based architectures \cite{iTransformer,PatchTST,crossformer,FEDformer} have become the dominant paradigm due to their ability to capture long-range dependencies and scalable global patterns across variables. More recently, a new line of work has emerged that leverages large language models \cite{touvron2023llama,bert} for time-series forecasting, exploring prompt-based reprogramming \cite{TEMPO}, cross-modal alignment strategies \cite{TimeLLM,CALF}, and pretrained semantic embeddings \cite{GPT4TS}. Although LLM-driven approaches have achieved notable progress, aligning numerical time-series representations with the semantic structures of language models while preserving variable identity and avoiding spurious correlations remains challenging.
\subsection{Cross-Modal Alignment for Time Series}
\begin{figure*}[htbp]
    \vspace{-0.1in}
    \centering
    \includegraphics[width=0.75\linewidth]{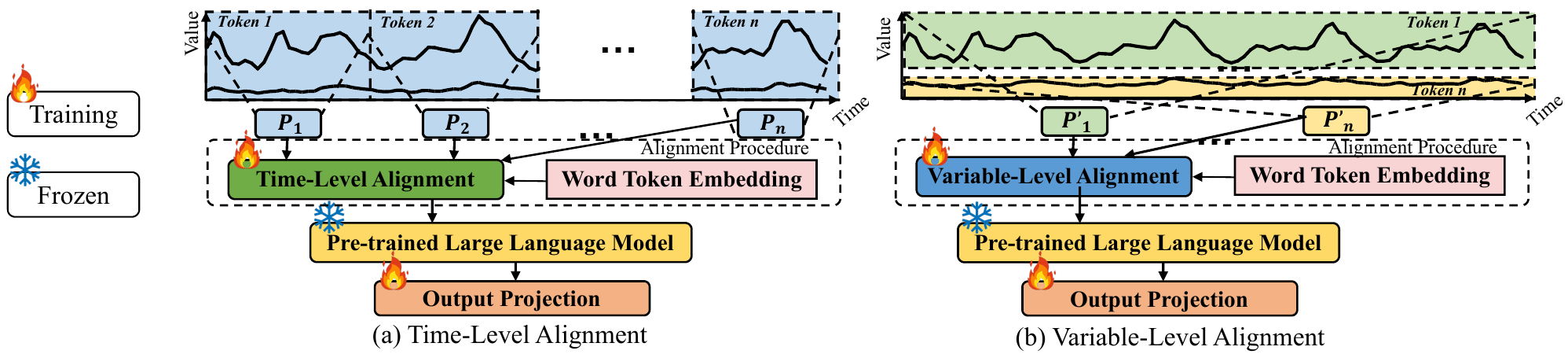}
    \caption{Cross-modal alignment is key to LLM-based time series models. Alignment strategies can typically be divided into (a) time-level and (b) variable-level approaches: the former treats each time step or patch as a token, while the latter considers the full sequence of a variable as a single token to preserve semantic consistency.}
    \label{related work f}
\end{figure*}

LLMs have shown strong generalization across domains such as NLP \cite{touvron2023llama}, vision \cite{visionllm}, and biology \cite{proteinllm}, prompting efforts to adapt them for time series forecasting \cite{PromptCast,GPT4TS}. However, the domain gap between natural language and numerical sequences poses significant challenges in LLM-based time series models. Early methods \cite{PromptCast,LLMTime} reformulate time series as text, enabling LLM usage but failing to capture intrinsic temporal structures. GPT4TS \cite{GPT4TS} and TimeLLM \cite{TimeLLM} address \textcolor{blue}{these} limitations via cross-modal alignment, mapping time series into the pretrained word embedding space. As summarized in Fig. \ref{related work f}, alignment strategies fall into time-level and variable-level approaches. Time-level alignment maps individual time steps or patches to language tokens, using feature mapping \cite{TEMPO,GPT4TS}, attention \cite{Autotimes,TimeLLM}, or graphs \cite{DECA}. Variable-level alignment, as in CALF \cite{CALF}, treats entire variable as tokens for cross-modal alignment. However, CALF \cite{CALF} may overlook the confounding effects: without disentangling dynamic and invariant components, the alignment can suffer from spurious correlations induced by the dynamic components, hence degrading representation quality.

\section{Method} \label{method}
In time series forecasting, given a multivariate series $\boldsymbol{X}\in \mathbb{R}^{T\times D}$ of length $T$ with $D$ variables, the goal is to learn a function $\mathcal{F}(\cdot)$ that predicts the next $S$ steps: $\boldsymbol{X}_{t-T+1:t}\xrightarrow{\mathcal{F}(\cdot)}\boldsymbol{\hat Y}_{t+1:t+S}$, where $\boldsymbol{\hat Y}\in\mathbb{R}^{S\times D}$ denotes the forecasted outputs over the next $S$ time steps. When $\mathcal{F}(\cdot)$ is instantiated with LLMs, strong performance requires bridging the domain gap between numerical time-series representations and the linguistic embedding space \cite{CALF}, while preserving temporal dynamics and semantic interpretability.

\begin{figure*}[h]
    \centering
    \includegraphics[width=0.8\linewidth]{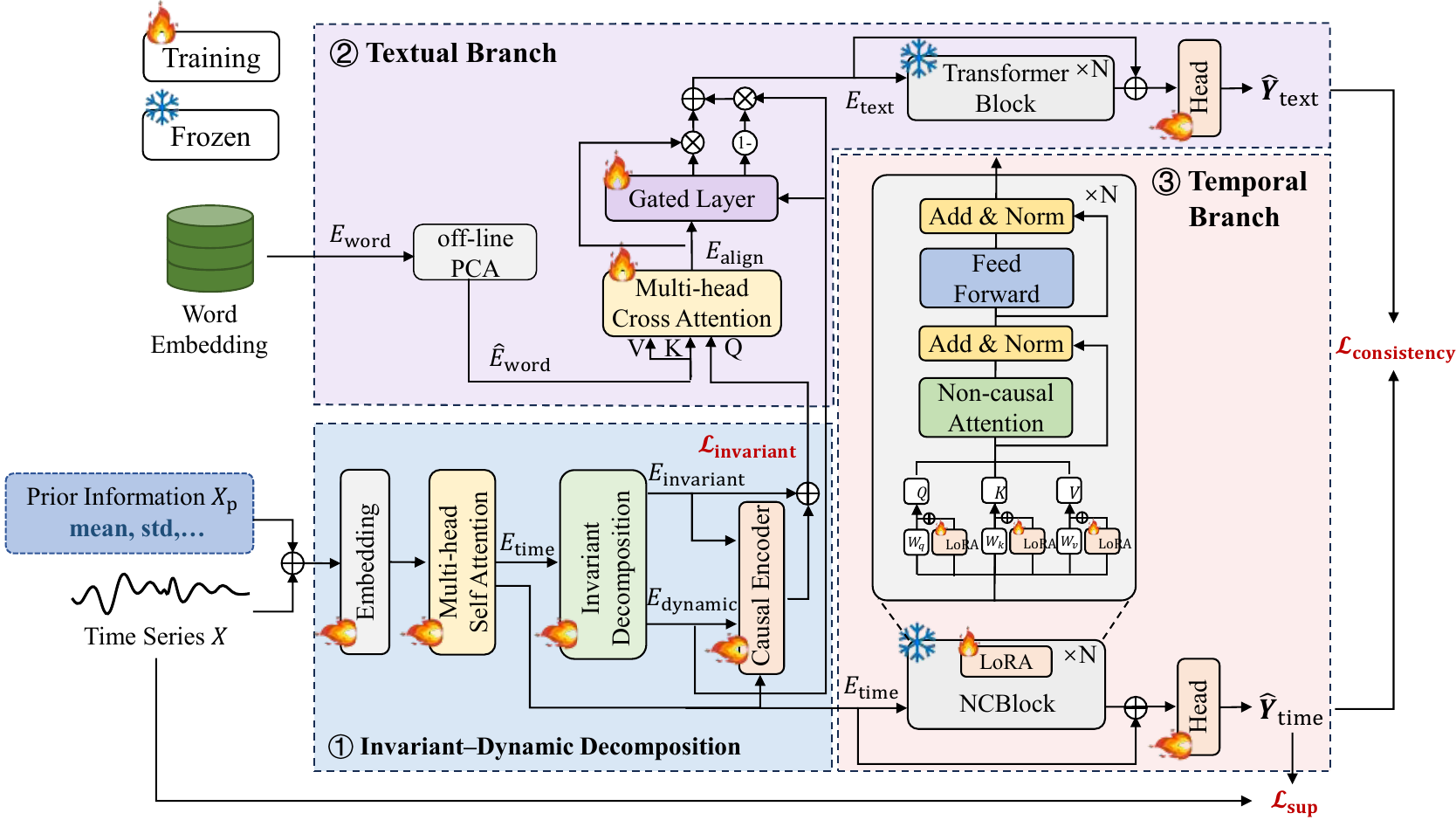}
    \caption{The framework of CVAformer, which performs alignment at the variable level. \ding{172} \textbf{Invariant–Dynamic Decomposition} embeds the input time series and separates each variable into an invariant semantic component and a dynamic component. 
\ding{173} \textbf{Textual Branch} conducts causally guided variable-level alignment by incorporating word embeddings, cross-attention, and a gated intervention mechanism to produce $\hat{Y}_{\text{text}}$. 
\ding{174} \textbf{Temporal Branch} models unordered inter-variable dependencies using stacked LLM blocks with non-causal attention and outputs $\hat{Y}_{\text{time}}$. 
Supervision and invariance losses ($\mathcal{L}_{\text{sup}}$ and $\mathcal{L}_{\text{invariant}}$) train the decomposition and temporal forecasting components, while a consistency loss $\mathcal{L}_{\text{consistency}}$ encourages agreement between the two branches.}
    \label{structure}
\end{figure*}

As shown in Fig. \ref{structure}, CVAformer processes inputs through three stages. The model first embeds the input series and decomposes each variable into invariant and dynamic representations (shown in \ding{172}). These representations then enter into the textual branch (shown in \ding{173}) and the temporal branch (shown in \ding{174}). The textual branch performs causally guided variable-level alignment by conditioning the invariant representation on a causal summary of the dynamics (e.g., global statistics and covariance patterns that characterize the current environmental regime) and matching it to pretrained semantic prototypes, producing $\hat{Y}_{\text{text}}$. In parallel, the temporal branch applies stacked LLM blocks with non-causal attention to capture unordered inter-variable dependencies and outputs $\hat{Y}_{\text{time}}$. 
\subsection{Semantic Grounding for Variable-Level Alignment}
Traditional forecasting models often process each time step independently, overlooking the rich inter-variable relationships in multivariate time series. As noted in \cite{iTransformer}, tokenization based solely on time points results in narrow receptive fields and misaligned semantics across variables. To address this, CVAformer adopts a novel variable-level alignment strategy that explicitly maps each variable to a semantic prototype derived from pretrained word embeddings.

Given multivariate time series $\boldsymbol{X}$, an embedding layer followed by Multi-head Self Attention (MHSA) \cite{iTransformer,CALF} is applied to extract time-aware representations $E_{\text{time}}$. To incorporate contextual knowledge, auxiliary prior information $\boldsymbol{X}_{\text{p}}$ is concatenated:
\begin{equation}
    E_\text{time} = \text{MHSA}(\text{Embedding}(\boldsymbol{X}||\boldsymbol{X}_{\text{p}}))\in \mathbb{R}^{\textcolor{black}{D\times H}},
\end{equation}
where $H$ is the hidden dimension of the pretrained LLM and $||$ denotes the concatenation operator. Prior information $\boldsymbol{X}_{\text{p}}$ contains simple global statistics computed from each variable over the processing window, including mean, standard deviation, and a linear trend slope coefficient. The embedding layer $\text{Embedding}(\cdot)$ projects each variable’s temporal window from $T$ to $H$ dimensions. To establish semantic grounding, a pretrained word embedding matrix $E_{\text{word}} \in \mathbb{R}^{V \times D}$ \textcolor{black}{is} extracted from the token embedding layer of the pretrained language model, where each row encodes the semantic representation of a word token learned during large-scale pretraining and $V$ denotes the vocabulary size. Directly performing cross-attention between time-series representations and all $V$ word vectors incurs substantial computational overhead and introduces semantic redundancy, as many words occupy tightly clustered regions in the embedding space. Leveraging this structure, we construct a compact semantic basis by applying off-line Principal Component Analysis (PCA) \cite{TimeLLM, CALF} to the pretrained word-embedding matrix ${E}_{\text{word}}$. PCA identifies the dominant axes of variance that correspond to coherent semantic neighborhoods, \textcolor{black}{thereby condensing the extensive vocabulary into a compact set of representative prototypes.}\\
This yields a reduced prototype matrix $\hat{E}_{\text{word}} = \text{PCA}(E_{\text{word}}) \in \mathbb{R}^{d \times D}$ with $d \ll V$, where each prototype captures a principal semantic direction shared across a group of related words. Importantly, this extraction is performed once prior to training, thereby adding no learnable parameters and preserving the semantic structure already encoded in the pretrained embedding. The resulting prototype set $\hat{E}_{\text{word}}$ is used exclusively in the textual branch for variable-level alignment, while the shared time representation $E_{\text{time}}$ serves both branches.
\subsection{Temporal Branch for Inter-Variable Dependencies}
The temporal branch is designed to model temporal dependencies among variables. Most LLM-based time series models adopt attention mechanisms to capture temporal dependencies \cite{TimeLLM, GPT4TS, Autotimes}, and typically employ causal masking to enforce autoregressive behavior. This is implemented using an additive causal mask $M$ that restricts attention to past and current positions:
\begin{equation}
M_{i,j}=\begin{cases} 0, & \text{if } j \leq i \\ -\infty, & \text{if } j > i \end{cases},
\end{equation}
where ${M}_{i,j}$ specifies whether the query at time step $i$ is permitted to attend to the key at time step $j$. As the mask is applied before the softmax operation, assigning ${M}_{i,j}=-\infty$ guarantees a strict zero attention weight for all future positions under softmax normalization. However, it is not suitable for capturing variable-wise interactions, which are unordered. Introducing an artificial temporal ordering across variables as channel-wise embeddings with implicit ordering \cite{CALF}, might impose inductive biases misaligned with the nature of multivariate time-series interactions.

To overcome this limitation, CVAformer employs stacked Non-Causal Attention Blocks (NCBlocks) that remove causal masking and allow each variable to attend to all variables within the same time step without imposing any ordering constraints. As illustrated in Fig. \ref{structure}, each NCBlock comprises a non-causal multi-head attention module, feed-forward layers, and residual connections with normalization.

Let $E_{\text{time}} \in \mathbb{R}^{D \times H}$ denote the embeddings of $D$ variables at a specific time step. The attention mechanism is computed as:
\begin{equation}
\text{Attention}(Q, K, V) = \text{softmax}\left(\frac{QK^\top}{\sqrt{H}}\right)V,
\end{equation}
where $Q = E_{\text{time}}W_Q$, $K = E_{\text{time}}W_K$, and $V = E_{\text{time}}W_V$ are the query, key, and value obtained via linear projections. Since no positional or causal masks are applied, all variables can mutually participate, enabling the model to learn relational dependencies independent of input ordering.

Multiple NCBlocks are stacked to capture higher-order variable dependencies while maintaining permutation equivariance across variable dimensions, ensuring consistent performance regardless of channel ordering. The enhanced $E_{\text{time}}$ is finally projected to the forecast $\boldsymbol{\hat{Y}}_{\text{time}}$ through residual connections and a prediction head.

\subsection{Text Branch for Alignment}
The textual branch performs variable-level alignment between $E_{\text{time}}$ and $\hat{E}_{\text{word}}$, reducing the representational mismatch between numerical time-series embeddings and pretrained semantic embeddings. Previous work \cite{CALF} leverages multi-head cross-attention for this purpose, similar to many time-level alignment methods \cite{TimeLLM}. However, these approaches often overlook the confounding effect of dynamic fluctuations, leading to spurious correlations during alignment by semantic drift, as illustrated in Fig. \ref{motivation} (b).

\subsubsection{Causal Lens for Alignment}
Existing LLM-based time series models typically align the directly extracted time embedding with the pre-trained word embedding of a language model. However, such alignment operates at the token level and ignores the fact that each variable contains both an invariant semantic identity and a dynamic component influenced by external conditions. When these factors are entangled, aligning the full time embedding can introduce spurious semantic correlations and produce unstable or inconsistent mappings. To address this limitation, our method performs variable-level alignment that explicitly targets the invariant semantic component of each variable. As illustrated in Fig. \ref{motivation} (a), the ideal alignment path is $I \rightarrow A$, where $I$ denotes the invariant semantics and $A$ denotes the aligned embedding. In practice, the dynamic component $C$ influences the estimation of the invariant part i.e., $C\rightarrow I$, causing the learned invariant representation to drift. Moreover, $C$ also implicitly affects the aligned embedding through the alignment procedure as $C\rightarrow A$. Together, these dependencies induce the backdoor path $I \leftarrow C \rightarrow A$ which introduces biased semantic alignment and may suppress the information that should truly reflect variable identity.\\
To mitigate this confounding effect and recover an unbiased alignment, we adopt a causal intervention based on the backdoor adjustment principle \cite{runge2023causal}. 
Specifically, under the causal graph in Fig. \ref{motivation} (a), the interventional distribution can be derived as:
\begin{align}
    & P(A \mid \operatorname{do}(I)) \\
    = &\sum_{c} P(A \mid \operatorname{do}(I), C=c)\, P(C=c \mid \operatorname{do}(I)) \label{cau_eq1}\\
    = &\sum_{c} P(A \mid \operatorname{do}(I), C=c)\, P(C=c) \label{cau_eq2}\\
    = &\sum_{c} P(A \mid I, C=c)\, P(C=c), \label{cau_eq}
\end{align}
where Eq. (\ref{cau_eq1}) follows from the law of total probability, Eq. (\ref{cau_eq2}) holds because the intervention on $I$ does not affect the confounder $C$ in Fig. \ref{motivation} (a), and Eq. (\ref{cau_eq}) replaces intervention with observation since conditioning on $C$ blocks the backdoor path between $I$ and $A$ in Fig. \ref{motivation} (a). The $\operatorname{do}(\cdot)$ operator removes the incoming edges to $I$, thereby blocking the confounding path via $C$.
\subsubsection{Decomposition}
\begin{figure}[htbp]
    \centering
    \includegraphics[width=0.8\linewidth]{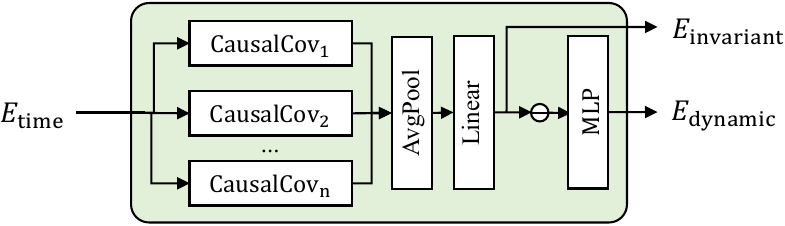}
    \caption{The procedure of decomposition for invariant and dynamic features.}
    \label{decomposition}
\end{figure}

We model the time embedding $E_{\text{time}}$ as a mixture of two sources: (i) a stable semantic component $E_{\text{invariant}}$, corresponding to $I$ in Fig. \ref{motivation} (a), and (ii) a dynamic component $E_{\text{dynamic}}$ that reflects short-term fluctuations induced by external conditions, corresponding to the confounder $C$ in Fig. \ref{motivation} (a). The overall decomposition is illustrated in Fig. \ref{decomposition}.

Multiple CausalCov blocks are applied to capture temporal dependencies across diverse receptive fields and provide $E_{\text{invariant}}$. Each block computes local covariance representations with their outputs aggregated via average pooling. With the learned $E_{\text{invariant}}$, $E_{\text{dynamic}}$ is then inferred with a learnable MLP layer. It treats the residual between the observed embedding and its invariant estimate as a proxy for external dynamics, while the MLP provides a flexible parametrization that can adaptively reshape the dynamic features.

To facilitate a more effective disentanglement between invariant and dynamic components, a contrastive loss inspired by the Momentum Contrast (MoCo) idea \cite{MoCo,woo2022cost} is employed. The key idea is that the invariant representation of each variable should remain stable under temporal augmentations, whereas the dynamic component is expected to change. The contrastive objective therefore encourages $E_{\text{invariant}}$ to be consistent across augmented views and prevents it from being influenced by dynamic variations $E_{\text{dynamic}}$, serving as a regularizer that suppresses entanglement between the two components. Implementation details are provided in the Appendix.

Given a batch of $N$ samples and a memory queue of $K_{\text{negative}}$ negative keys, the contrastive objective is formulated as:
\begin{equation}
    \mathcal{L}_{\text{invariant}} = \sum_{i=1}^{N}-\text{log}\frac{\text{exp}(q_{i}\cdot k_{i}/\tau)}{\text{exp}(q_{i}\cdot k_{i}/\tau)+\sum_{j=1}^{K_{\text{negative}}}\text{exp}(q_{i}\cdot k_{j}/\tau)},
\end{equation}
where $q_{i}$ is obtained by applying a projection head on $E^{i}_{\text{invariant}^{}}$, $k_{i}$ is its positive augmented counterpart, and $\tau$ is a temperature hyperparameter. Negative keys $k_{j}$ are sampled from a cross-batch queue.

This design encourages the model to emphasize stable global patterns in $E_{\text{invariant}}$ while allowing $E_{\text{invariant}}$ to encode fine-grained temporal fluctuations. In subsequent alignment modules, $E_{\text{dynamic}}$ is treated as a potential confounder when performing approximate backdoor adjustment.

\subsubsection{Causal Alignment}
To perform semantic alignment under changing environmental conditions, we incorporate causal adjustment into the alignment module. Rather than conditioning directly on the raw dynamic embedding $E_{\text{dynamic}}$, which is often high-dimensional and sensitive to local fluctuations, we extract a compact confounder summary $E_{\text{causal}}$ that characterizes the global context of environmental variation. This representation is designed to approximate the marginal confounder distribution $P(C)$ required for the backdoor adjustment in Eq. \ref{cau_eq}.

Given the decomposed representations $E_{\text{invariant}}$, $E_{\text{dynamic}}$, and the full temporal embedding $E_{\text{time}}$, we first compute the temporal covariance matrix:
\begin{equation}
\hat{E}_{\text{cov}} = \text{Cov}(E_{\text{time}}) = \frac{1}{T - 1} E_{\text{time}} \cdot E_{\text{time}}^{T},
\end{equation}
where $\hat E_{\text{cov}}$ denotes a covariance-style second-order statistic summarizing cross-variable temporal dynamics. To summarize the global dynamic condition, we obtain the temporal means $\overline{E}_{\text{dynamic}}$ and $\overline{E}_{\text{invariant}}$, and the covariance $\hat{E}_{\text{cov}}$. These components capture short-term fluctuations, stable semantic trends, and cross-variable temporal dependencies, respectively. They are concatenated and fed into a lightweight two-layer network with GELU activation and LayerNorm:
\begin{equation}
E_{\text{causal}} = \text{CausalEncoder}(\overline{E}_{\text{invariant}} || \overline{E}_{\text{dynamic}} || \hat{E}_{\text{cov}}) \in \mathbb{R}^{\textcolor{black}{D \times H}}.
\end{equation}
The resulting $E_{\text{causal}}$ serves as a compact proxy for the confounder context, providing the model with a stable conditioning signal that mitigates the influence of short-term environmental fluctuations.

After disentanglement and confounder adjustment, we perform alignment using Multi-Head Cross Attention (MHCA). To integrate causal information into the alignment process, the query is constructed by combining the invariant representation with the confounder summary:
\begin{align}
   & E_{\text{align}} = \text{MHCA}(Q,K,V),\\
   & Q=(E_{\text{invariant}}+E_{\text{causal}})W_{q},\\ 
   & K=\hat{E}_{\text{word}}W_{k}, \\ 
   & V=\hat{E}_{\text{word}}W_{v},
\end{align}
where $W_{q}, W_{k}, W_{v}$ are learnable projection matrices. This design ensures that alignment is guided primarily by stable semantics while being properly conditioned on the underlying environment, effectively approximating $P(A \mid I, C)$.

To flexibly integrate the aligned semantic features with environment-driven variations, we further introduce a variable-wise gating mechanism:
\begin{align}
    &G = \text{GateLayer}(E_{\text{align}}||E_{\text{dynamic}}),\\
    &E_{\text{text}}=G\cdot E_{\text{align}}+(1-G)\cdot E_{\text{dynamic}},
\end{align}
where $G \in (0,1)^{H \times D}$ controls the relative contribution of invariant-driven semantic alignment and dynamic fluctuations for each variable. This gated fusion yields the final causally adjusted textual representation $E_{\text{text}}$, which is then projected to $\hat{\boldsymbol{Y}}_{\text{text}}$ through residual connections and a prediction head.

\subsection{Optimization Objective}
In addition to $\mathcal{L}_{\text{invariant}}$, which stabilizes the decomposition of each variable into invariant and dynamic components, two additional objectives are required to properly guide the learning process. The supervised loss $\mathcal{L}_{\text{sup}}$ ensures that the temporal branch is directly optimized for forecasting accuracy, anchoring the model to the predictive task. The consistency loss $\mathcal{L}_{\text{consistency}}$ enforces semantic agreement between the textual and temporal branches, preventing them from converging toward incompatible representations and ensuring that both branches encode coherent variable-level semantics. Together, these objectives complement one another: $\mathcal{L}_{\text{invariant}}$ enhances semantic disentanglement, $\mathcal{L}_{\text{sup}}$ ensures task grounding, and $\mathcal{L}_{\text{consistency}}$ maintains cross-branch coherence.

In detail, $\mathcal{L}_{\text{sup}}$ is computed between the temporal prediction $\boldsymbol{\hat Y}_{\text{time}}$ and ground truth. Given the output of temporal branches $\boldsymbol{\hat Y}_{\text{time}}$ and textual branches $\boldsymbol{\hat Y}_{\text{text}}$, the consistency loss is defined as
\begin{equation}
    \mathcal{L}_{\text{consistency}}=\text{sim}(\boldsymbol{\hat Y}_{\text{text}},\boldsymbol{\hat Y}_{\text{time}}),
\end{equation}
where $\text{sim}(\cdot,\cdot)$ is the similarity function as $L_{1}$ loss. The total loss $\mathcal{L}_{\text{total}}$ to be used during training is the weighted summation of $\mathcal{L}_{\text{sup}}$, $\mathcal{L}_{\text{consistency}}$, and $\mathcal{L}_{\text{invariant}}$ as:
\begin{equation}
\mathcal{L}_{\text{total}} = \mathcal{L}_{\text{sup}} + \lambda_{1} \mathcal{L}_{\text{consistency}} + \lambda_{2} \mathcal{L}_{\text{invariant}},
\end{equation}
where $\lambda_{1}$ and $\lambda_{2}$ are hyperparameters. 

To support efficient optimization within the modular architecture of CVAformer, we adopt a lightweight finetuning strategy that separates structural adaptation from semantic grounding. The pretrained LLM backbone already provides rich semantic priors, and fully updating its parameters would be computationally expensive and increase the risk of overfitting on limited time-series data. For this reason, the feed-forward layers of the backbone remain frozen, while only a small set of adaptable components: positional embeddings, normalization layers, and LoRA adapters \cite{Lora}, are updated during training.

LoRA \cite{Lora} is applied to the attention projection matrices so that the model can adjust its relational reasoning capacity to the temporal dependencies present in time-series data while keeping the backbone parameters intact. This finetuning strategy integrates naturally with the overall structure of CVAformer by preserving the pretrained semantic space and providing an efficient mechanism for task-specific adaptation.

\section{Experiments}\label{experiment}
To comprehensively evaluate the effectiveness of the proposed CVAformer, a series of experiments are conducted across four tasks: long-term forecasting, short-term forecasting, few-shot forecasting, and zero-shot forecasting.

\textbf{Baselines.} Several representative models from time series community are selected, including the following categories: (1) LLM-based models: TimeLLM \cite{TimeLLM}, GPT4TS \cite{GPT4TS}, and CALF \cite{CALF}; (2) Transformer-based models: iTransformer \cite{iTransformer}, PatchTST \cite{PatchTST}, FEDformer \cite{FEDformer}, Crossformer \cite{crossformer}; (3) MLP-based models: DLinear \cite{DLinear}; (4) CNN-based models: TimesNet \cite{TimesNet}. For short-term forecasting tasks, additional MLP-based models as N-HiTS \cite{Nhits} and N-BEATS \cite{Nbeats} are included due to their strong performance in this setting. All models are implemented under the protocol established by TimesNet \cite{TimesNet}. The pretrained GPT-2 model \cite{radford2019language} is used as the default backbone LLM in CVAformer. The details of the implementation can be found in Appendix \ref{Experimental-appendix}.

\subsection{Long-term forecasting}
\textbf{Setups.} Seven widely used datasets are employed for evaluation, including Weather\footnote{\url{https://ncei.noaa.gov/data/local-climatological-data}\label{weafoot}}, Traffic\footnote{\url{http://pems.dot.ca.gov/}}, Electricity\footnote{\url{https://archive.ics.uci.edu/ml/datasets/ElectricityLoadDiagrams20112014}\label{elefoot}}, and four ETT datasets \cite{informer} (ETTh1, ETTh2, ETTm1, ETTm2). To ensure fair comparison, the input sequence length is fixed to 96, and the prediction horizons are set to $\{96, 192, 336, 720\}$. Mean Square Error (MSE) and Mean Absolute Error (MAE) are adopted as evaluation metrics.

\textbf{Results.} The average forecasting results are reported in Table \ref{long-term}, with the best results \textbf{bold} and the second best \underline{underlined}. CVAformer consistently outperforms all baseline methods on long-term forecasting tasks, demonstrating strong capability in capturing long-range temporal dependencies through causality-aware modeling. Notably, it achieves relative reductions of 10.07\% in MSE and 8.21\% in MAE compared to PatchTST \cite{PatchTST}. Compared with the LLM-based method GPT4TS \cite{GPT4TS}, CVAformer yields an average reduction of 6.82\% in MSE and 5.71\% in MAE. A notable exception occurs on the Traffic dataset, where iTransformer attains slightly better results. This dataset is dominated by highly regular periodic patterns, which align well with iTransformer's strong channel-mixing inductive bias. In contrast, CVAformer's semantic and causal modeling brings greater benefits to datasets with richer cross-variable interactions. Even so, CVAformer remains competitive on Traffic and achieves the best overall results across the full set of long-term forecasting benchmarks.
\subsection{Short-term forecasting}
\textbf{Setups.} The M4 dataset \cite{M4}, which contains yearly, quarterly, and monthly univariate marketing time series, is used for short-term forecasting. The prediction horizons in this case are relatively small as $[6, 48]$, while the input lengths are set to twice the corresponding horizons. The evaluation metrics include Symmetric Mean Absolute Percentage Error (SMAPE), Mean Absolute Scaled Error (MASE), and Overall Weighted Average (OWA).

\textbf{Results.} The results are presented in Table \ref{short-term}. CVAformer has demonstrated its capacity in short-term forecasting across various evaluation metrics, ranking first in 11 out of 15 evaluation cases. Compared to PatchTST \cite{PatchTST}, it yields an average improvement of 2.8\%, and achieves a 2.2\% overall reduction compared to GPT4TS \cite{GPT4TS}. For the remaining cases where CVAformer is not the best performer, the differences are generally small. These cases typically correspond to very short forecasting horizons, where fine-grained local patterns play a dominant role and can be captured more effectively by architectures with strong local inductive biases. CVAformer, by contrast, emphasizes variable-level semantic alignment and causal disentanglement, which yield larger benefits in settings that require richer semantic structure. Despite these few exceptions, CVAformer remains highly competitive across all tasks and achieves the best overall performance.
\setlength{\tabcolsep}{3.5pt}
\begin{table*}[htbp]
\centering
\caption{Comparison of various models on long-term forecasting tasks with different datasets. The input sequence length is 96 for all baselines.}
\label{long-term}
\tiny
\begin{tabular}{cccccccccccccccccccccc}
\toprule
\multicolumn{2}{c}{Models}  & \multicolumn{2}{c}{{\makecell[c]{CVAformer\\ (Ours)}}}& \multicolumn{2}{c}{{\makecell[c]{TimeLLM\\ \cite{TimeLLM}}}}& \multicolumn{2}{c}{\makecell[c]{GPT4TS\\ \cite{GPT4TS}}}    & \multicolumn{2}{c}{\makecell[c]{CALF\\ \cite{CALF}}}& \multicolumn{2}{c}{\makecell[c]{iTransformer\\ \cite{iTransformer}}}& \multicolumn{2}{c}{\makecell[c]{PatchTST\\ \cite{PatchTST}}}&\multicolumn{2}{c}{\makecell[c]{FEDformer\\ \cite{FEDformer}}}  &\multicolumn{2}{c}{\makecell[c]{Crossformer\\ \cite{crossformer}}}  &\multicolumn{2}{c}{\makecell[c]{DLinear\\ \cite{DLinear}}}  &\multicolumn{2}{c}{\makecell[c]{TimesNet\\ \cite{TimesNet}}} \\ \midrule
\multicolumn{2}{c}{Metrics} & MSE                  & \multicolumn{1}{c}{MAE} & MSE                  & \multicolumn{1}{c}{MAE} & MSE                  & \multicolumn{1}{c}{MAE} & MSE                  & \multicolumn{1}{c}{MAE} & MSE                  & \multicolumn{1}{c}{MAE} & MSE                  & \multicolumn{1}{c}{MAE} &  MSE                  & \multicolumn{1}{c}{MAE}  &  MSE                  & \multicolumn{1}{c}{MAE}  &  MSE                  & \multicolumn{1}{c}{MAE}  &  MSE                  & \multicolumn{1}{c}{MAE}  \\ 
\midrule
\multicolumn{1}{c}{\multirow{5}{*}{\rotatebox{90}{ETTh1}}}&\multicolumn{1}{c}{96}&\textbf{0.368$\pm$0.005}& \multicolumn{1}{c}{\textbf{0.388$\pm$0.001}}&0.398& \multicolumn{1}{c}{0.410}&\underline{0.376}& \multicolumn{1}{c}{0.397}&{\underline{0.376}}& \multicolumn{1}{c}{0.393}&0.394& \multicolumn{1}{c}{0.409}&0.377& \multicolumn{1}{c}{0.396}&\underline{0.376}& \multicolumn{1}{c}{0.419}&0.420& \multicolumn{1}{c}{0.439}&0.386& \multicolumn{1}{c}{0.402}&0.384& \multicolumn{1}{c}{0.402}\\
\multicolumn{1}{c}{}&\multicolumn{1}{c}{192}&\underline{0.421$\pm$0.007}&\multicolumn{1}{c}{\textbf{0.421$\pm$0.010}}&0.451&\multicolumn{1}{c}{0.440}&0.438&\multicolumn{1}{c}{0.426}&0.430&\multicolumn{1}{c}{0.427}&0.448&\multicolumn{1}{c}{0.440}&0.425&\multicolumn{1}{c}{0.426}&\textbf{0.420}&\multicolumn{1}{c}{0.448}&0.541&\multicolumn{1}{c}{0.520}&0.437&\multicolumn{1}{c}{0.432}&0.436&0.429\\
\multicolumn{1}{c}{}&\multicolumn{1}{c}{336}&0.463$\pm$0.004&\multicolumn{1}{c}{\underline{0.444$\pm$0.007}}&0.508&\multicolumn{1}{c}{0.471}&0.479&\multicolumn{1}{c}{0.446}&0.482&\multicolumn{1}{c}{0.452}&0.492&\multicolumn{1}{c}{0.465}&0.461&\multicolumn{1}{c}{0.448}&\textbf{0.459}&\multicolumn{1}{c}{0.465}&0.722&\multicolumn{1}{c}{0.648}&0.481&\multicolumn{1}{c}{0.459}&0.491&0.469\\
\multicolumn{1}{r}{}&\multicolumn{1}{c}{720}&\textbf{0.462$\pm$0.011}&\multicolumn{1}{c}{\underline{0.462$\pm$0.012}}&0.483&\multicolumn{1}{c}{0.478}&0.495&\multicolumn{1}{c}{0.476}&0.485&\multicolumn{1}{c}{0.471}&0.520&\multicolumn{1}{c}{0.503}&0.529&\multicolumn{1}{c}{0.500}&0.506&\multicolumn{1}{c}{0.507}&0.811&\multicolumn{1}{c}{0.691}&0.519&\multicolumn{1}{c}{0.516}&0.521&0.500\\
\multicolumn{1}{r}{}&\multicolumn{1}{c}{Avg.}&\textbf{0.428}&\multicolumn{1}{c}{\textbf{0.428}}&0.460&\multicolumn{1}{c}{0.449}&0.447&\multicolumn{1}{c}{0.436}&0.443&\multicolumn{1}{c}{0.435}&0.464&\multicolumn{1}{c}{0.454}&0.448&\multicolumn{1}{c}{0.443}&\underline{0.440}&\multicolumn{1}{c}{0.460}&0.623&\multicolumn{1}{c}{0.574}&0.456&\multicolumn{1}{c}{0.452}&0.458&0.450\\
\midrule
\multicolumn{1}{c}{\multirow{5}{*}{\rotatebox{90}{ETTh2}}}&\multicolumn{1}{c}{96}&\textbf{0.281$\pm$0.005}& \multicolumn{1}{c}{\textbf{0.331$\pm$0.004}}&0.295& \multicolumn{1}{c}{0.346}&0.295& \multicolumn{1}{c}{0.348}&\underline{0.290}& \multicolumn{1}{c}{\underline{0.337}}&0.297& \multicolumn{1}{c}{0.349}&0.310& \multicolumn{1}{c}{0.353}&0.358& \multicolumn{1}{c}{0.397}&0.745& \multicolumn{1}{c}{0.584}&0.333& \multicolumn{1}{c}{0.387}&0.340& \multicolumn{1}{c}{0.374}\\
\multicolumn{1}{r}{}&\multicolumn{1}{c}{192}&\textbf{0.360$\pm$0.002}&\multicolumn{1}{c}{\textbf{0.379$\pm$0.006}}&0.386&\multicolumn{1}{c}{0.399}&0.386&\multicolumn{1}{c}{0.404}&\underline{0.367}&\multicolumn{1}{c}{\underline{0.385}}&0.380&\multicolumn{1}{c}{0.400}&0.390&\multicolumn{1}{c}{0.405}&0.429&\multicolumn{1}{c}{0.439}&0.877&\multicolumn{1}{c}{0.656}&0.477&\multicolumn{1}{c}{0.476}&0.402&0.414\\
\multicolumn{1}{r}{}&\multicolumn{1}{c}{336}&\textbf{0.407$\pm$0.005}&\multicolumn{1}{c}{\textbf{0.419$\pm$0.009}}&0.447&\multicolumn{1}{c}{0.443}&0.421&\multicolumn{1}{c}{0.435}&\underline{0.415}&\multicolumn{1}{c}{\underline{0.422}}&0.428&\multicolumn{1}{c}{0.432}&0.430&\multicolumn{1}{c}{0.434}&0.496&\multicolumn{1}{c}{0.487}&1.043&\multicolumn{1}{c}{0.731}&0.594&\multicolumn{1}{c}{0.541}&0.452&0.452\\
\multicolumn{1}{r}{}&\multicolumn{1}{c}{720}&\textbf{0.412$\pm$0.001}&\multicolumn{1}{c}{{\textbf{0.424$\pm$0.003}}}&{0.428}&\multicolumn{1}{c}{0.444}&0.422&\multicolumn{1}{c}{0.445}&\underline{0.415}&\multicolumn{1}{c}{\underline{0.435}}&0.427&\multicolumn{1}{c}{0.445}&0.438&\multicolumn{1}{c}{0.449}&0.463&\multicolumn{1}{c}{0.474}&1.104&\multicolumn{1}{c}{0.763}&0.831&\multicolumn{1}{c}{0.657}&0.462&0.468\\
\multicolumn{1}{r}{}&\multicolumn{1}{c}{Avg.}&\textbf{0.365}&\multicolumn{1}{c}{\textbf{0.388}}&0.389&\multicolumn{1}{c}{0.408}&0.381&\multicolumn{1}{c}{0.408}&\underline{0.371}&\multicolumn{1}{c}{\underline{0.394}}&0.383&\multicolumn{1}{c}{0.407}&0.392&\multicolumn{1}{c}{0.410}&0.437&\multicolumn{1}{c}{0.449}&0.942&\multicolumn{1}{c}{0.684}&0.559&\multicolumn{1}{c}{0.515}&0.414&0.427\\
\midrule
\multicolumn{1}{c}{\multirow{5}{*}{\rotatebox{90}{ETTm1}}}&\multicolumn{1}{c}{96}&\textbf{0.311$\pm$0.003}& \multicolumn{1}{c}{\textbf{0.337$\pm$0.002}}&0.359& \multicolumn{1}{c}{0.381}&0.329& \multicolumn{1}{c}{0.364}&\underline{0.323}& \multicolumn{1}{c}{\underline{0.349}}&0.343& \multicolumn{1}{c}{0.378}&0.332& \multicolumn{1}{c}{0.369}&0.379& \multicolumn{1}{c}{0.419}&0.370& \multicolumn{1}{c}{0.404}&0.345& \multicolumn{1}{c}{0.372}&0.338& \multicolumn{1}{c}{0.375}\\
\multicolumn{1}{r}{}&\multicolumn{1}{c}{192}&\textbf{0.362$\pm$0.001}&\multicolumn{1}{c}{\textbf{0.383$\pm$0.003}}&0.383&\multicolumn{1}{c}{0.393}&0.368&\multicolumn{1}{c}{0.382}&0.376&\multicolumn{1}{c}{\underline{0.376}}&0.381&\multicolumn{1}{c}{0.395}&\underline{0.373}&\multicolumn{1}{c}{0.389}&0.426&\multicolumn{1}{c}{0.441}&0.460&\multicolumn{1}{c}{0.488}&0.380&\multicolumn{1}{c}{0.389}&0.374&0.387\\
\multicolumn{1}{r}{}&\multicolumn{1}{c}{336}&\textbf{0.402$\pm$0.005}&\multicolumn{1}{c}{\textbf{0.396$\pm$0.004}}&0.416&\multicolumn{1}{c}{0.414}&0.400&\multicolumn{1}{c}{0.403}&0.410&\multicolumn{1}{c}{\underline{0.399}}&0.419&\multicolumn{1}{c}{0.418}&\underline{0.408}&\multicolumn{1}{c}{0.411}&0.445&\multicolumn{1}{c}{0.459}&0.637&\multicolumn{1}{c}{0.607}&0.413&\multicolumn{1}{c}{0.413}&0.410&0.411\\
\multicolumn{1}{r}{}&\multicolumn{1}{c}{720}&\textbf{0.463$\pm$0.001}&\multicolumn{1}{c}{\textbf{0.432$\pm$0.003}}&0.483&\multicolumn{1}{c}{0.449}&0.460&\multicolumn{1}{c}{0.439}&0.476&\multicolumn{1}{c}{\underline{0.436}}&0.486&\multicolumn{1}{c}{0.456}&\underline{0.463}&\multicolumn{1}{c}{0.446}&0.543&\multicolumn{1}{c}{0.490}&0.863&\multicolumn{1}{c}{0.720}&0.474&\multicolumn{1}{c}{0.453}&0.478&0.450\\
\multicolumn{1}{r}{}&\multicolumn{1}{c}{Avg.}&\textbf{0.384}&\multicolumn{1}{c}{\textbf{0.387}}&0.410&\multicolumn{1}{c}{0.409}&0.389&\multicolumn{1}{c}{0.397}&0.394&\multicolumn{1}{c}{\underline{0.389}}&0.407&\multicolumn{1}{c}{0.411}&\underline{0.394}&\multicolumn{1}{c}{0.404}&0.448&\multicolumn{1}{c}{0.452}&0.582&\multicolumn{1}{c}{0.555}&0.403&\multicolumn{1}{c}{0.407}&0.400&0.406\\
\midrule
\multicolumn{1}{c}{\multirow{5}{*}{\rotatebox{90}{ETTm2}}}&\multicolumn{1}{c}{96}&\textbf{0.171$\pm$0.009}& \multicolumn{1}{c}{\textbf{0.249$\pm$0.007}}&0.193& \multicolumn{1}{c}{0.280}&0.178& \multicolumn{1}{c}{0.263}&{0.178}& \multicolumn{1}{c}{\underline{0.256}}&0.183& \multicolumn{1}{c}{0.268}&\underline{0.177}& \multicolumn{1}{c}{0.259}&0.203& \multicolumn{1}{c}{0.287}&0.270& \multicolumn{1}{c}{0.372}&0.193& \multicolumn{1}{c}{0.292}&0.187& \multicolumn{1}{c}{0.267}\\
\multicolumn{1}{r}{}&\multicolumn{1}{c}{192}&\textbf{0.239$\pm$0.003}&\multicolumn{1}{c}{\textbf{0.295$\pm$0.002}}&0.257&\multicolumn{1}{c}{0.318}&0.245&\multicolumn{1}{c}{0.306}&\underline{0.242}&\multicolumn{1}{c}{\underline{0.297}}&0.252&\multicolumn{1}{c}{0.312}&\underline{0.242}&\multicolumn{1}{c}{0.301}&0.269&\multicolumn{1}{c}{0.328}&0.345&\multicolumn{1}{c}{0.401}&0.284&\multicolumn{1}{c}{0.362}&0.249&0.309\\
\multicolumn{1}{r}{}&\multicolumn{1}{c}{336}&\textbf{0.302$\pm$0.002}&\multicolumn{1}{c}{\textbf{0.336$\pm$0.002}}&0.317&\multicolumn{1}{c}{0.353}&0.309&\multicolumn{1}{c}{0.347}&{0.307}&\multicolumn{1}{c}{\underline{0.339}}&0.313&\multicolumn{1}{c}{0.350}&\underline{0.303}&\multicolumn{1}{c}{0.343}&0.325&\multicolumn{1}{c}{0.366}&0.650&\multicolumn{1}{c}{0.524}&0.369&\multicolumn{1}{c}{0.427}&0.321&0.351\\
\multicolumn{1}{r}{}&\multicolumn{1}{c}{720}&\textbf{0.389$\pm$0.005}&\multicolumn{1}{c}{\textbf{0.384$\pm$0.006}}&0.419&\multicolumn{1}{c}{0.411}&0.409&\multicolumn{1}{c}{0.408}&\underline{0.397}&\multicolumn{1}{c}{\underline{0.393}}&0.411&\multicolumn{1}{c}{0.406}&0.410&\multicolumn{1}{c}{0.404}&0.421&\multicolumn{1}{c}{0.415}&1.208&\multicolumn{1}{c}{0.753}&0.554&\multicolumn{1}{c}{0.522}&0.408&0.403\\
\multicolumn{1}{r}{}&\multicolumn{1}{c}{Avg.}&\textbf{0.275}&\multicolumn{1}{c}{\textbf{0.316}}&0.296&\multicolumn{1}{c}{0.340}&0.285&\multicolumn{1}{c}{0.331}&\underline{0.281}&\multicolumn{1}{c}{\underline{0.321}}&0.290&\multicolumn{1}{c}{0.334}&{0.283}&\multicolumn{1}{c}{0.327}&0.305&\multicolumn{1}{c}{0.349}&0.618&\multicolumn{1}{c}{0.513}&0.350&\multicolumn{1}{c}{0.401}&0.291&0.333\\
\midrule
\multicolumn{1}{c}{\multirow{5}{*}{\rotatebox{90}{Weather}}}&\multicolumn{1}{c}{96}&\textbf{0.160$\pm$0.002}& \multicolumn{1}{c}{\textbf{0.202$\pm$0.005}}&0.195& \multicolumn{1}{c}{0.233}&0.182& \multicolumn{1}{c}{0.223}&\underline{0.166}& \multicolumn{1}{c}{\underline{0.206}}&0.174& \multicolumn{1}{c}{0.214}&0.179& \multicolumn{1}{c}{0.220}&0.217& \multicolumn{1}{c}{0.296}&0.234& \multicolumn{1}{c}{0.305}&0.196& \multicolumn{1}{c}{0.255}&0.172& \multicolumn{1}{c}{0.220}\\
\multicolumn{1}{r}{}&\multicolumn{1}{c}{192}&\textbf{0.207$\pm$0.004}&\multicolumn{1}{c}{\textbf{0.247$\pm$0.004}}&0.240&\multicolumn{1}{c}{0.269}&0.231&\multicolumn{1}{c}{0.263}&\underline{0.215}&\multicolumn{1}{c}{\underline{0.251}}&0.221&\multicolumn{1}{c}{0.254}&0.222&\multicolumn{1}{c}{0.257}&0.276&\multicolumn{1}{c}{0.336}&0.368&\multicolumn{1}{c}{0.407}&0.237&\multicolumn{1}{c}{0.296}&0.219&0.261\\
\multicolumn{1}{r}{}&\multicolumn{1}{c}{336}&\textbf{0.265$\pm$0.005}&\multicolumn{1}{c}{\textbf{0.287$\pm$0.002}}&0.293&\multicolumn{1}{c}{0.306}&0.283&\multicolumn{1}{c}{0.300}&\underline{0.269}&\multicolumn{1}{c}{\underline{0.291}}&0.278&\multicolumn{1}{c}{0.296}&0.279&\multicolumn{1}{c}{0.297}&0.339&\multicolumn{1}{c}{0.380}&0.490&\multicolumn{1}{c}{0.510}&0.283&\multicolumn{1}{c}{0.335}&0.280&0.306\\
\multicolumn{1}{r}{}&\multicolumn{1}{c}{720}&\textbf{0.348$\pm$0.005}&\multicolumn{1}{c}{\textbf{0.346$\pm$0.007}}&0.368&\multicolumn{1}{c}{0.354}&0.360&\multicolumn{1}{c}{0.350}&\underline{0.355}&\multicolumn{1}{c}{\underline{0.352}}&0.358&\multicolumn{1}{c}{0.349}&0.357&\multicolumn{1}{c}{0.351}&0.403&\multicolumn{1}{c}{0.428}&0.525&\multicolumn{1}{c}{0.543}&0.345&\multicolumn{1}{c}{0.381}&0.365&0.359\\
\multicolumn{1}{r}{}&\multicolumn{1}{c}{Avg.}&\textbf{0.245}&\multicolumn{1}{c}{\textbf{0.270}}&0.274&\multicolumn{1}{c}{0.290}&0.264&\multicolumn{1}{c}{0.284}&\underline{0.250}&\multicolumn{1}{c}{\underline{0.274}}&0.258&\multicolumn{1}{c}{0.279}&0.259&\multicolumn{1}{c}{0.281}&0.309&\multicolumn{1}{c}{0.360}&0.404&\multicolumn{1}{c}{0.441}&0.265&\multicolumn{1}{c}{0.317}&0.259&0.287\\
\midrule
\multicolumn{1}{c}{\multirow{5}{*}{\rotatebox{90}{Electricity}}}&\multicolumn{1}{c}{96}&\textbf{0.140$\pm$0.008}& \multicolumn{1}{c}{\textbf{0.231$\pm$0.05}}&0.204& \multicolumn{1}{c}{0.293}&0.185& \multicolumn{1}{c}{0.272}&\underline{0.145}& \multicolumn{1}{c}{\underline{0.238}}&0.148& \multicolumn{1}{c}{0.240}&0.170& \multicolumn{1}{c}{0.260}&0.193& \multicolumn{1}{c}{0.308}&0.150& \multicolumn{1}{c}{0.253}&0.197& \multicolumn{1}{c}{0.282}&0.168& \multicolumn{1}{c}{0.272}\\
\multicolumn{1}{r}{}&\multicolumn{1}{c}{192}&\textbf{0.160$\pm$0.001}&\multicolumn{1}{c}{\textbf{0.252$\pm$0.003}}&0.207&\multicolumn{1}{c}{0.295}&0.189&\multicolumn{1}{c}{0.276}&\underline{0.161}&\multicolumn{1}{c}{\textbf{0.252}}&0.162&\multicolumn{1}{c}{\underline{0.253}}&0.187&\multicolumn{1}{c}{0.276}&0.201&\multicolumn{1}{c}{0.315}&0.167&\multicolumn{1}{c}{0.267}&0.196&\multicolumn{1}{c}{0.285}&0.184&0.289\\
\multicolumn{1}{r}{}&\multicolumn{1}{c}{336}&\textbf{0.170$\pm$0.005}&\multicolumn{1}{c}{\textbf{0.264$\pm$0.009}}&0.219&\multicolumn{1}{c}{0.308}&0.204&\multicolumn{1}{c}{0.291}&\underline{0.175}&\multicolumn{1}{c}{\underline{0.267}}&0.178&\multicolumn{1}{c}{0.269}&0.203&\multicolumn{1}{c}{0.291}&0.214&\multicolumn{1}{c}{0.329}&0.189&\multicolumn{1}{c}{0.287}&0.209&\multicolumn{1}{c}{0.301}&0.198&0.300\\
\multicolumn{1}{r}{}&\multicolumn{1}{c}{720}&\textbf{0.205$\pm$0.003}&\multicolumn{1}{c}{\textbf{0.294$\pm$0.006}}&0.263&\multicolumn{1}{c}{0.341}&0.245&\multicolumn{1}{c}{0.324}&\underline{0.222}&\multicolumn{1}{c}{\underline{0.303}}&0.225&\multicolumn{1}{c}{0.317}&0.245&\multicolumn{1}{c}{0.325}&0.246&\multicolumn{1}{c}{0.355}&0.256&\multicolumn{1}{c}{0.337}&0.245&\multicolumn{1}{c}{0.333}&0.220&0.320\\
\multicolumn{1}{r}{}&\multicolumn{1}{c}{Avg.}&\textbf{0.168}&\multicolumn{1}{c}{\textbf{0.260}}&0.223&\multicolumn{1}{c}{0.309}&0.205&\multicolumn{1}{c}{0.290}&\underline{0.175}&\multicolumn{1}{c}{\underline{0.265}}&0.178&\multicolumn{1}{c}{0.270}&0.201&\multicolumn{1}{c}{0.288}&0.214&\multicolumn{1}{c}{0.327}&0.191&\multicolumn{1}{c}{0.286}&0.212&\multicolumn{1}{c}{0.300}&0.192&0.295\\
\midrule
\multicolumn{1}{c}{\multirow{5}{*}{\rotatebox{90}{Traffic}}}&\multicolumn{1}{c}{96}&\underline{0.401$\pm$0.004}& \multicolumn{1}{c}{\textbf{0.263$\pm$0.005}}&0.536& \multicolumn{1}{c}{0.359}&0.468& \multicolumn{1}{c}{0.307}&{0.408}& \multicolumn{1}{c}{{0.269}}&\textbf{0.395}& \multicolumn{1}{c}{\underline{0.268}}&0.544& \multicolumn{1}{c}{0.360}&0.587& \multicolumn{1}{c}{0.366}&0.522& \multicolumn{1}{c}{0.290}&0.650& \multicolumn{1}{c}{0.396}&0.593& \multicolumn{1}{c}{0.321}\\
\multicolumn{1}{r}{}&\multicolumn{1}{c}{192}&\underline{0.424$\pm$0.005}&\multicolumn{1}{c}{\textbf{0.274$\pm$0.003}}&0.530&\multicolumn{1}{c}{0.354}&0.476&\multicolumn{1}{c}{0.311}&{0.430}&\multicolumn{1}{c}{{0.278}}&\textbf{0.417}&\multicolumn{1}{c}{\underline{0.276}}&0.540&\multicolumn{1}{c}{0.354}&0.604&\multicolumn{1}{c}{0.373}&0.530&\multicolumn{1}{c}{0.293}&0.598&\multicolumn{1}{c}{0.370}&0.617&0.336\\
\multicolumn{1}{r}{}&\multicolumn{1}{c}{336}&\underline{0.444$\pm$0.006}&\multicolumn{1}{c}{\textbf{0.276$\pm$0.005}}&0.530&\multicolumn{1}{c}{0.349}&0.488&\multicolumn{1}{c}{0.317}&{0.446}&\multicolumn{1}{c}{{0.283}}&\textbf{0.433}&\multicolumn{1}{c}{\underline{0.283}}&0.552&\multicolumn{1}{c}{0.360}&0.621&\multicolumn{1}{c}{0.383}&0.558&\multicolumn{1}{c}{0.305}&0.605&\multicolumn{1}{c}{0.373}&0.629&0.336\\
\multicolumn{1}{r}{}&\multicolumn{1}{c}{720}&\textbf{0.465$\pm$0.009}&\multicolumn{1}{c}{\textbf{0.289$\pm$0.009}}&0.569&\multicolumn{1}{c}{0.371}&0.521&\multicolumn{1}{c}{0.333}&{0.477}&\multicolumn{1}{c}{\underline{0.300}}&\underline{0.467}&\multicolumn{1}{c}{0.302}&0.590&\multicolumn{1}{c}{0.378}&0.626&\multicolumn{1}{c}{0.382}&0.589&\multicolumn{1}{c}{0.328}&0.645&\multicolumn{1}{c}{0.394}&0.640&0.350\\
\multicolumn{1}{r}{}&\multicolumn{1}{c}{Avg.}&\textbf{0.433}&\multicolumn{1}{c}{\textbf{0.275}}&0.541&\multicolumn{1}{c}{0.358}&0.488&\multicolumn{1}{c}{0.317}&\underline{0.439}&\multicolumn{1}{c}{\underline{0.281}}&0.428&\multicolumn{1}{c}{0.282}&0.557&\multicolumn{1}{c}{0.363}&0.610&\multicolumn{1}{c}{0.376}&0.550&\multicolumn{1}{c}{0.304}&0.625&\multicolumn{1}{c}{0.383}&0.620&0.336\\
\bottomrule
\end{tabular}
\vspace{-0.1in}
\end{table*}
\begin{table*}[!htbp]
\centering
\caption{Comparison of various models on short-term forecasting tasks with M4 dataset. The input sequence length are $[12, 96]$ and the prediction length are $[6, 48]$ for all baselines. }
\label{short-term}
\tiny
\begin{tabular}{cccccccccccccc}
\toprule
\multicolumn{2}{c}{Models}  & \multicolumn{1}{c}{{\makecell[c]{CVAformer\\ (Ours)}}}& \multicolumn{1}{c}{\makecell[c]{TimeLLM\\ \cite{TimeLLM}}}& \multicolumn{1}{c}{\makecell[c]{GPT4TS\\ \cite{GPT4TS}}}    & \multicolumn{1}{c}{\makecell[c]{CALF\\ \cite{CALF}}}& \multicolumn{1}{c}{\makecell[c]{iTransformer\\ \cite{iTransformer}}}& \multicolumn{1}{c}{\makecell[c]{PatchTST\\ \cite{PatchTST}}}&\multicolumn{1}{c}{\makecell[c]{FEDformer\\ \cite{FEDformer}}}  &\multicolumn{1}{c}{\makecell[c]{Crossformer\\ \cite{crossformer}}}  &\multicolumn{1}{c}{\makecell[c]{DLinear\\ \cite{DLinear}}}  &\multicolumn{1}{c}{\makecell[c]{TimesNet\\ \cite{TimesNet}}} &\multicolumn{1}{c}{\makecell[c]{N-HiTS\\ \cite{Nhits}}}& \multicolumn{1}{c}{\makecell[c]{N-BEATS\\ \cite{Nbeats}}}\\
\midrule
\multicolumn{1}{r}{\multirow{3}{*}{{\rotatebox{90}{Yearly}}}}&\multicolumn{1}{c}{SMAPE}&\textbf{13.183$\pm$0.109}&13.419&13.531&\underline{13.351}&14.321&13.550&13.728&{13.392}&16.965&13.387&13.418&13.436\\
\multicolumn{1}{r}{}&\multicolumn{1}{c}{MASE}&\textbf{2.946$\pm$0.015}&3.005&3.015&3.003&3.230&3.280&3.048&{3.001}&4.283&\underline{2.996}&3.045&3.043\\
\multicolumn{1}{r}{}&\multicolumn{1}{c}{OWA}&\textbf{0.774$\pm$0.009}&0.789&0.793&{0.786}&0.845&0.796&0.803&{0.787}&\underline{0.781}&{0.786}&0.793&0.794\\
\midrule
\multicolumn{1}{r}{\multirow{3}{*}{{\rotatebox{90}{Quarterly}}}}&\multicolumn{1}{c}{SMAPE}&\underline{9.995$\pm$0.112}&10.110&10.177&\textbf{9.990}&10.764&10.195&10.792&16.317&12.145&10.100&10.202&10.124\\
\multicolumn{1}{r}{}&\multicolumn{1}{c}{MASE}&\textbf{1.161$\pm$0.007}&1.178&1.194&\underline{1.164}&1.284&1.205&1.283&2.197&1.520&1.182&1.194&{1.169}\\
\multicolumn{1}{r}{}&\multicolumn{1}{c}{OWA}&\textbf{0.877$\pm$0.003}&0.889&0.898&\underline{0.878}&0.956&0.902&0.958&1.542&1.106&0.890&0.899&0.886\\
\midrule
\multicolumn{1}{r}{\multirow{3}{*}{{\rotatebox{90}{Monthly}}}}&\multicolumn{1}{c}{SMAPE}&\textbf{12.632$\pm$0.105}&12.980&12.894&\underline{12.643}&14.245&12.960&14.260&12.924&13.514&12.670&12.791&{12.677}\\
\multicolumn{1}{r}{}&\multicolumn{1}{c}{MASE}&\textbf{0.922$\pm$0.007}&0.963&0.956&\textbf{0.922}&1.110&0.968&1.102&0.966&1.037&\underline{0.933}&0.969&{0.937}\\
\multicolumn{1}{r}{}&\multicolumn{1}{c}{OWA}&\textbf{0.871$\pm$0.005}&0.903&0.897&\underline{0.872}&1.016&0.905&1.012&0.902&0.956&{0.878}&0.899&{0.880}\\
\midrule
\multicolumn{1}{r}{\multirow{3}{*}{{\rotatebox{90}{Others}}}}&\multicolumn{1}{c}{SMAPE}&\underline{4.618$\pm$0.111}&4.795&4.940&\textbf{4.552}&5.780&5.005&4.954&5.493&6.709&4.891&5.061&4.925\\
\multicolumn{1}{r}{}&\multicolumn{1}{c}{MASE}&\underline{3.165$\pm$0.013}&3.178&3.228&\textbf{3.092}&4.153&3.394&3.264&3.690&4.953&3.302&3.216&3.391\\
\multicolumn{1}{r}{}&\multicolumn{1}{c}{OWA}&\underline{0.985$\pm$0.010}&1.006&1.029&\textbf{0.967}&1.263&1.062&1.036&1.160&1.487&1.035&1.040&1.053\\
\midrule
\multicolumn{1}{r}{\multirow{3}{*}{{\rotatebox{90}{Average}}}}&\multicolumn{1}{c}{SMAPE}&\textbf{11.725}&\underline{11.765}&11.991&\underline{11.765}&12.999&12.034&12.840&13.474&13.639&{11.829}&11.927&11.851\\
\multicolumn{1}{r}{}&\multicolumn{1}{c}{MASE}&\textbf{1.558}&\underline{1.567}&1.595&\underline{1.567}&1.792&1.620&1.701&1.866&2.095&{1.585}&1.613&1.599\\
\multicolumn{1}{r}{}&\multicolumn{1}{c}{OWA}&\textbf{0.840}&\underline{0.844}&0.859&\underline{0.844}&0.948&0.867&0.918&0.985&1.051&{0.851}&0.861&0.855\\
\bottomrule
\end{tabular}
\vspace{-0.1in}
\end{table*}
\setlength{\tabcolsep}{1pt}
\subsection{Few/Zero-shot Forecasting}
\textbf{Setups.} To evaluate the robustness and generalization capability of CVAformer under limited-data conditions, we conduct few-shot and zero-shot forecasting experiments on the ETT dataset \cite{informer}.

\textbf{Few-shot Forecasting.}
In the few-shot forecasting setting, only a small portion of the training data is made available for model learning. Specifically, the first 10\% of the original training split is used for each dataset, creating a data-constrained training regime. The results are summarized in Table \ref{few-shot}. CVAformer consistently outperforms both deep learning-based and LLM-based time series models, achieving an average error reduction of 13\% compared to PatchTST \cite{PatchTST}, and 10.3\% compared to GPT4TS \cite{GPT4TS}.

\textbf{Zero-shot Forecasting.} In the zero-shot forecasting setting, the model is trained on a source domain $\clubsuit$ and directly evaluated on a target domain $\spadesuit$ that remains unseen during training. As shown in Table \ref{zero-shot}, CVAformer achieves consistent gains over both LLM-based and deep learning-based baselines. Notably, CVAformer demonstrates strong cross-domain generalization, outperforming PatchTST \cite{PatchTST} and GPT4TS \cite{GPT4TS} by 17.7\% and 11.2\% on average, respectively.

Although CVAformer shows a few cases where it does not achieve the top score in zero-shot and few-shot settings, the differences are small and do not alter the overall performance trend. The model remains competitive across all datasets.
\setlength{\tabcolsep}{4pt}
\setlength{\tabcolsep}{3.5pt}
\begin{table*}[htbp]
\centering
\caption{Comparison of various models on few-shot forecasting tasks with 10\% training data on ETT datasets. The input sequence length is 96 for all baselines.}
\label{few-shot}
\tiny
\begin{tabular}{cccccccccccccccccccccc}
\toprule
\multicolumn{2}{c}{Models}  & \multicolumn{2}{c}{{\makecell[c]{CVAformer\\ (Ours)}}}& \multicolumn{2}{c}{{\makecell[c]{TimeLLM\\ \cite{TimeLLM}}}}& \multicolumn{2}{c}{\makecell[c]{GPT4TS\\ \cite{GPT4TS}}}    & \multicolumn{2}{c}{\makecell[c]{CALF\\ \cite{CALF}}}& \multicolumn{2}{c}{\makecell[c]{iTransformer\\ \cite{iTransformer}}}& \multicolumn{2}{c}{\makecell[c]{PatchTST\\ \cite{PatchTST}}}&\multicolumn{2}{c}{\makecell[c]{FEDformer\\ \cite{FEDformer}}}  &\multicolumn{2}{c}{\makecell[c]{Crossformer\\ \cite{crossformer}}}  &\multicolumn{2}{c}{\makecell[c]{DLinear\\ \cite{DLinear}}}  &\multicolumn{2}{c}{\makecell[c]{TimesNet\\ \cite{TimesNet}}} \\ \midrule
\multicolumn{2}{c}{Metrics} & MSE                  & \multicolumn{1}{c}{MAE} & MSE                  & \multicolumn{1}{c}{MAE} & MSE                  & \multicolumn{1}{c}{MAE} & MSE                  & \multicolumn{1}{c}{MAE} & MSE                  & \multicolumn{1}{c}{MAE} & MSE                  & \multicolumn{1}{c}{MAE} &  MSE                  & \multicolumn{1}{c}{MAE}  &  MSE                  & \multicolumn{1}{c}{MAE}  &  MSE                  & \multicolumn{1}{c}{MAE}  &  MSE                  & \multicolumn{1}{c}{MAE}  \\ 
\midrule
\multicolumn{1}{c}{\multirow{5}{*}{\rotatebox{90}{ETTh1}}}&\multicolumn{1}{c}{96}&\underline{0.476$\pm$0.003}& \multicolumn{1}{c}{\underline{0.457$\pm$0.002}}&0.500& \multicolumn{1}{c}{0.464}&{0.462}& \multicolumn{1}{c}{{0.449}}&0.484& \multicolumn{1}{c}{0.463}&0.790& \multicolumn{1}{c}{0.586}&\textbf{0.433}& \multicolumn{1}{c}{\textbf{0.428}}&0.651& \multicolumn{1}{c}{0.563}&1.129& \multicolumn{1}{c}{0.775}&0.590& \multicolumn{1}{c}{0.515}&0.855& \multicolumn{1}{c}{0.625}\\
\multicolumn{1}{r}{}&\multicolumn{1}{c}{192}&\textbf{0.533$\pm$0.005}&\multicolumn{1}{c}{\textbf{0.501$\pm$0.004}}&0.590&\multicolumn{1}{c}{0.516}&0.551&\multicolumn{1}{c}{0.495}&\underline{0.535}&\multicolumn{1}{c}{\underline{0.490}}&0.837&\multicolumn{1}{c}{0.609}&\underline{0.509}&\multicolumn{1}{c}{\underline{0.474}}&0.666&\multicolumn{1}{c}{0.562}&1.832&\multicolumn{1}{c}{0.922}&0.634&\multicolumn{1}{c}{0.541}&0.791&0.589\\
\multicolumn{1}{r}{}&\multicolumn{1}{c}{336}&\textbf{0.620$\pm$0.002}&\multicolumn{1}{c}{\textbf{0.537$\pm$0.001}}&0.638&\multicolumn{1}{c}{0.542}&0.630&\multicolumn{1}{c}{0.539}&0.631&\multicolumn{1}{c}{0.541}&0.780&\multicolumn{1}{c}{0.575}&\underline{0.572}&\multicolumn{1}{c}{\underline{0.509}}&0.767&\multicolumn{1}{c}{0.602}&2.022&\multicolumn{1}{c}{0.973}&0.659&\multicolumn{1}{c}{0.554}&0.939&0.648\\
\multicolumn{1}{r}{}&\multicolumn{1}{c}{720}&\textbf{0.987$\pm$0.003}&\multicolumn{1}{c}{\textbf{0.688$\pm$0.002}}&1.334&\multicolumn{1}{c}{0.816}&1.113&\multicolumn{1}{c}{0.738}&\underline{0.998}&\multicolumn{1}{c}{\underline{0.692}}&1.234&\multicolumn{1}{c}{0.811}&1.221&\multicolumn{1}{c}{0.773}&{0.918}&\multicolumn{1}{c}{{0.703}}&1.903&\multicolumn{1}{c}{0.986}&0.708&\multicolumn{1}{c}{0.598}&0.876&0.641\\
\multicolumn{1}{r}{}&\multicolumn{1}{c}{Avg.}&\textbf{0.653}&\multicolumn{1}{c}{\textbf{0.545}}&0.765&\multicolumn{1}{c}{0.584}&0.689&\multicolumn{1}{c}{0.555}&\underline{0.662}&\multicolumn{1}{c}{\underline{0.546}}&0.910&\multicolumn{1}{c}{0.860}&0.683&\multicolumn{1}{c}{0.645}&0.750&\multicolumn{1}{c}{0.607}&1.744&\multicolumn{1}{c}{0.914}&0.647&\multicolumn{1}{c}{0.552}&0.865&0.625\\
\midrule
\multicolumn{1}{c}{\multirow{5}{*}{\rotatebox{90}{ETTh2}}}&\multicolumn{1}{c}{96}&\textbf{0.312$\pm$0.002}& \multicolumn{1}{c}{\textbf{0.363$\pm$0.004}}&0.329& \multicolumn{1}{c}{0.365}&0.327& \multicolumn{1}{c}{0.359}&\underline{0.323}& \multicolumn{1}{c}{\underline{0.362}}&0.404& \multicolumn{1}{c}{0.435}&{0.314}& \multicolumn{1}{c}{{0.354}}&0.359& \multicolumn{1}{c}{0.404}&2.482& \multicolumn{1}{c}{1.206}&0.361& \multicolumn{1}{c}{0.407}&0.372& \multicolumn{1}{c}{0.405}\\
\multicolumn{1}{r}{}&\multicolumn{1}{c}{192}&\underline{0.412$\pm$0.005}&\multicolumn{1}{c}{\underline{0.413$\pm$0.003}}&0.414&\multicolumn{1}{c}{0.413}&\textbf{0.403}&\multicolumn{1}{c}{\textbf{0.405}}&0.431&\multicolumn{1}{c}{0.422}&0.470&\multicolumn{1}{c}{0.474}&{0.420}&\multicolumn{1}{c}{{0.415}}&0.460&\multicolumn{1}{c}{0.461}&3.136&\multicolumn{1}{c}{1.372}&0.444&\multicolumn{1}{c}{0.453}&0.488&0.463\\
\multicolumn{1}{r}{}&\multicolumn{1}{c}{336}&\textbf{0.512$\pm$0.003}&\multicolumn{1}{c}{\textbf{0.481$\pm$0.001}}&0.579&\multicolumn{1}{c}{0.506}&0.568&\multicolumn{1}{c}{0.499}&0.583&\multicolumn{1}{c}{0.512}&0.489&\multicolumn{1}{c}{0.485}&\underline{0.543}&\multicolumn{1}{c}{\underline{0.489}}&0.569&\multicolumn{1}{c}{0.530}&2.925&\multicolumn{1}{c}{1.331}&0.509&\multicolumn{1}{c}{0.501}&0.541&0.496\\
\multicolumn{1}{r}{}&\multicolumn{1}{c}{720}&\textbf{0.518$\pm$0.003}&\multicolumn{1}{c}{\textbf{0.495$\pm$0.002}}&1.034&\multicolumn{1}{c}{0.711}&1.020&\multicolumn{1}{c}{0.725}&\underline{0.521}&\multicolumn{1}{c}{\underline{0.498}}&0.593&\multicolumn{1}{c}{0.538}&0.926&\multicolumn{1}{c}{0.691}&0.827&\multicolumn{1}{c}{0.707}&4.014&\multicolumn{1}{c}{1.603}&0.453&\multicolumn{1}{c}{0.471}&0.510&0.491\\
\multicolumn{1}{r}{}&\multicolumn{1}{c}{Avg.}&\textbf{0.438}&\multicolumn{1}{c}{\textbf{0.444}}&0.589&\multicolumn{1}{c}{0.498}&0.579&\multicolumn{1}{c}{0.497}&\underline{0.464}&\multicolumn{1}{c}{\underline{0.448}}&0.489&\multicolumn{1}{c}{0.483}&0.550&\multicolumn{1}{c}{0.487}&0.553&\multicolumn{1}{c}{0.525}&3.139&\multicolumn{1}{c}{1.378}&0.441&\multicolumn{1}{c}{0.458}&0.476&0.463\\
\midrule
\multicolumn{1}{c}{\multirow{5}{*}{\rotatebox{90}{ETTm1}}}&\multicolumn{1}{c}{96}&\textbf{0.451$\pm$0.002}& \multicolumn{1}{c}{\textbf{0.435$\pm$0.002}}&0.587& \multicolumn{1}{c}{0.491}&0.615& \multicolumn{1}{c}{0.497}&\underline{0.468}& \multicolumn{1}{c}{\underline{0.445}}&0.709& \multicolumn{1}{c}{0.556}&0.558& \multicolumn{1}{c}{0.478}&0.604& \multicolumn{1}{c}{0.530}&1.037& \multicolumn{1}{c}{0.705}&0.552& \multicolumn{1}{c}{0.488}&0.583& \multicolumn{1}{c}{0.503}\\
\multicolumn{1}{r}{}&\multicolumn{1}{c}{192}&\textbf{0.463$\pm$0.004}&\multicolumn{1}{c}{\textbf{0.436$\pm$0.005}}&0.606&\multicolumn{1}{c}{0.490}&0.597&\multicolumn{1}{c}{0.492}&\underline{0.479}&\multicolumn{1}{c}{\underline{0.446}}&0.717&\multicolumn{1}{c}{0.548}&{0.539}&\multicolumn{1}{c}{{0.471}}&0.641&\multicolumn{1}{c}{0.546}&1.170&\multicolumn{1}{c}{0.778}&0.546&\multicolumn{1}{c}{0.487}&0.608&0.515\\
\multicolumn{1}{r}{}&\multicolumn{1}{c}{336}&\textbf{0.499$\pm$0.003}&\multicolumn{1}{c}{\textbf{0.463$\pm$0.003}}&0.719&\multicolumn{1}{c}{0.555}&0.597&\multicolumn{1}{c}{0.501}&\textbf{0.499}&\multicolumn{1}{c}{\textbf{0.463}}&0.735&\multicolumn{1}{c}{0.575}&{0.558}&\multicolumn{1}{c}{{0.488}}&0.768&\multicolumn{1}{c}{0.606}&1.463&\multicolumn{1}{c}{0.913}&0.567&\multicolumn{1}{c}{0.501}&0.733&0.572\\
\multicolumn{1}{r}{}&\multicolumn{1}{c}{720}&\textbf{0.568$\pm$0.002}&\multicolumn{1}{c}{\textbf{0.496$\pm$0.001}}&0.632&\multicolumn{1}{c}{0.514}&0.623&\multicolumn{1}{c}{0.513}&\underline{0.572}&\multicolumn{1}{c}{\underline{0.496}}&0.752&\multicolumn{1}{c}{0.584}&{0.574}&\multicolumn{1}{c}{{0.498}}&0.771&\multicolumn{1}{c}{0.606}&1.693&\multicolumn{1}{c}{0.997}&0.606&\multicolumn{1}{c}{0.522}&0.768&0.548\\
\multicolumn{1}{r}{}&\multicolumn{1}{c}{Avg.}&\textbf{0.495}&\multicolumn{1}{c}{\textbf{0.457}}&0.636&\multicolumn{1}{c}{0.512}&0.608&\multicolumn{1}{c}{0.500}&\underline{0.508}&\multicolumn{1}{c}{\underline{0.464}}&0.728&\multicolumn{1}{c}{0.565}&0.557&\multicolumn{1}{c}{0.483}&0.696&\multicolumn{1}{c}{0.572}&1.340&\multicolumn{1}{c}{0.848}&0.567&\multicolumn{1}{c}{0.499}&0.673&0.534\\
\midrule
\multicolumn{1}{c}{\multirow{5}{*}{\rotatebox{90}{ETTm2}}}&\multicolumn{1}{c}{96}&\textbf{0.187$\pm$0.001}& \multicolumn{1}{c}{\textbf{0.266$\pm$0.002}}&0.189& \multicolumn{1}{c}{0.270}&\textbf{0.187}& \multicolumn{1}{c}{\textbf{0.266}}&0.190& \multicolumn{1}{c}{0.268}&0.245& \multicolumn{1}{c}{0.322}&\underline{0.189}& \multicolumn{1}{c}{\underline{0.268}}&0.222& \multicolumn{1}{c}{0.314}&1.397& \multicolumn{1}{c}{0.866}&0.225& \multicolumn{1}{c}{0.320}&0.214& \multicolumn{1}{c}{0.288}\\
\multicolumn{1}{r}{}&\multicolumn{1}{c}{192}&\underline{0.253$\pm$0.002}&\multicolumn{1}{c}{\underline{0.308$\pm$0.001}}&0.264&\multicolumn{1}{c}{0.319}&\underline{0.253}&\multicolumn{1}{c}{\underline{0.308}}&0.257&\multicolumn{1}{c}{0.311}&0.274&\multicolumn{1}{c}{0.338}&\textbf{0.248}&\multicolumn{1}{c}{\textbf{0.307}}&0.284&\multicolumn{1}{c}{0.351}&1.757&\multicolumn{1}{c}{0.987}&0.291&\multicolumn{1}{c}{0.362}&0.271&0.325\\
\multicolumn{1}{r}{}&\multicolumn{1}{c}{336}&\textbf{0.320$\pm$0.004}&\multicolumn{1}{c}{\textbf{0.349$\pm$0.002}}&0.327&\multicolumn{1}{c}{0.358}&0.332&\multicolumn{1}{c}{0.353}&0.323&\multicolumn{1}{c}{0.334}&0.361&\multicolumn{1}{c}{0.394}&\underline{0.311}&\multicolumn{1}{c}{\underline{0.346}}&0.392&\multicolumn{1}{c}{0.419}&2.075&\multicolumn{1}{c}{1.086}&0.354&\multicolumn{1}{c}{0.402}&0.329&0.356\\
\multicolumn{1}{r}{}&\multicolumn{1}{c}{720}&{0.442$\pm$0.001}&\multicolumn{1}{c}{{0.419$\pm$0.003}}&0.454&\multicolumn{1}{c}{0.428}&\underline{0.438}&\multicolumn{1}{c}{\underline{0.417}}&0.441&\multicolumn{1}{c}{0.410}&0.467&\multicolumn{1}{c}{0.442}&\textbf{0.435}&\multicolumn{1}{c}{\textbf{0.418}}&0.527&\multicolumn{1}{c}{0.485}&2.712&\multicolumn{1}{c}{1.253}&0.446&\multicolumn{1}{c}{0.447}&0.473&0.448\\
\multicolumn{1}{r}{}&\multicolumn{1}{c}{Avg.}&\textbf{0.301}&\multicolumn{1}{c}{\textbf{0.335}}&0.308&\multicolumn{1}{c}{0.343}&{0.303}&\multicolumn{1}{c}{{0.336}}&\underline{0.302}&\multicolumn{1}{c}{\underline{0.330}}&0.336&\multicolumn{1}{c}{0.373}&{0.295}&\multicolumn{1}{c}{{0.334}}&0.356&\multicolumn{1}{c}{0.392}&1.985&\multicolumn{1}{c}{1.048}&0.329&\multicolumn{1}{c}{0.382}&0.321&0.354\\
\bottomrule
\end{tabular}
\vspace{-0.1in}
\end{table*}
\setlength{\tabcolsep}{2.5pt}
\begin{table*}[!htbp]
\centering
\caption{Comparison of various models on zero-shot forecasting tasks with different datasets, where 'h1', 'h2', 'm1', and 'm2' denote ETTh1, ETTh2, ETTm1, and ETTm2 respectively. $\clubsuit\xrightarrow{} \spadesuit$ indicates that models trained on the dataset $\clubsuit$ are evaluated on a distinct dataset $\spadesuit$. The input sequence length is 96 for all baselines.}
\label{zero-shot}
\tiny
\begin{tabular}{cccccccccccccccccccccc}
\toprule
\multicolumn{2}{c}{Models}  & \multicolumn{2}{c}{{\makecell[c]{CVAformer\\ (Ours)}}}& \multicolumn{2}{c}{{\makecell[c]{TimeLLM\\ \cite{TimeLLM}}}}& \multicolumn{2}{c}{\makecell[c]{GPT4TS\\ \cite{GPT4TS}}}    & \multicolumn{2}{c}{\makecell[c]{CALF\\ \cite{CALF}}}& \multicolumn{2}{c}{\makecell[c]{iTransformer\\ \cite{iTransformer}}}& \multicolumn{2}{c}{\makecell[c]{PatchTST\\ \cite{PatchTST}}}&\multicolumn{2}{c}{\makecell[c]{FEDformer\\ \cite{FEDformer}}}  &\multicolumn{2}{c}{\makecell[c]{Crossformer\\ \cite{crossformer}}}  &\multicolumn{2}{c}{\makecell[c]{DLinear\\ \cite{DLinear}}}  &\multicolumn{2}{c}{\makecell[c]{TimesNet\\ \cite{TimesNet}}} \\ \midrule
\multicolumn{2}{c}{Metrics} & MSE                  & \multicolumn{1}{c}{MAE} & MSE                  & \multicolumn{1}{c}{MAE} & MSE                  & \multicolumn{1}{c}{MAE} & MSE                  & \multicolumn{1}{c}{MAE} & MSE                  & \multicolumn{1}{c}{MAE} & MSE                  & \multicolumn{1}{c}{MAE} &  MSE                  & \multicolumn{1}{c}{MAE}  &  MSE                  & \multicolumn{1}{c}{MAE}  &  MSE                  & \multicolumn{1}{c}{MAE}  &  MSE                  & \multicolumn{1}{c}{MAE}  \\ 
\midrule
\multicolumn{1}{c}{\multirow{5}{*}{h1 $\xrightarrow{}$m1}}&\multicolumn{1}{c}{96}&\textbf{0.702$\pm$0.006}& \multicolumn{1}{c}{\textbf{0.547$\pm$0.005}}&0.804& \multicolumn{1}{c}{0.565}&0.809& \multicolumn{1}{c}{0.563}&0.809& \multicolumn{1}{c}{0.570}&1.336& \multicolumn{1}{c}{0.714}&0.908& \multicolumn{1}{c}{0.596}&\underline{0.731}& \multicolumn{1}{c}{\underline{0.561}}&0.856& \multicolumn{1}{c}{0.649}&{0.735}& \multicolumn{1}{c}{{0.554}}&0.764& \multicolumn{1}{c}{0.563}\\
\multicolumn{1}{r}{}&\multicolumn{1}{c}{192}&\textbf{0.711$\pm$0.004}&\multicolumn{1}{c}{\textbf{0.553$\pm$0.004}}&0.827&\multicolumn{1}{c}{0.593}&0.799&\multicolumn{1}{c}{0.567}&{0.759}&\multicolumn{1}{c}{{0.570}}&0.863&\multicolumn{1}{c}{0.590}&0.927&\multicolumn{1}{c}{0.616}&\underline{0.746}&\multicolumn{1}{c}{\underline{0.573}}&0.906&\multicolumn{1}{c}{0.684}&0.752&\multicolumn{1}{c}{0.570}&0.798&0.562\\
\multicolumn{1}{r}{}&\multicolumn{1}{c}{336}&\textbf{0.722$\pm$0.007}&\multicolumn{1}{c}{\textbf{0.563$\pm$0.005}}&0.835&\multicolumn{1}{c}{0.600}&0.803&\multicolumn{1}{c}{0.577}&\underline{0.746}&\multicolumn{1}{c}{\underline{0.575}}&1.032&\multicolumn{1}{c}{0.645}&0.920&\multicolumn{1}{c}{0.621}&0.775&\multicolumn{1}{c}{0.596}&1.104&\multicolumn{1}{c}{0.796}&0.749&\multicolumn{1}{c}{0.579}&0.790&0.584\\
\multicolumn{1}{r}{}&\multicolumn{1}{c}{720}&\underline{0.751$\pm$0.009}&\multicolumn{1}{c}{\textbf{0.582$\pm$0.006}}&0.922&\multicolumn{1}{c}{0.644}&\underline{0.783}&\multicolumn{1}{c}{\underline{0.589}}&0.760&\multicolumn{1}{c}{0.591}&0.954&\multicolumn{1}{c}{0.630}&{0.822}&\multicolumn{1}{c}{{0.608}}&0.808&\multicolumn{1}{c}{0.625}&1.131&\multicolumn{1}{c}{0.816}&0.805&\multicolumn{1}{c}{0.606}&0.827&0.594\\
\multicolumn{1}{r}{}&\multicolumn{1}{c}{Avg.}&\textbf{0.721}&\multicolumn{1}{c}{\textbf{0.561}}&0.847&\multicolumn{1}{c}{0.600}&0.798&\multicolumn{1}{c}{0.574}&\underline{0.768}&\multicolumn{1}{c}{\underline{0.576}}&1.046&\multicolumn{1}{c}{0.644}&0.894&\multicolumn{1}{c}{0.610}&0.765&\multicolumn{1}{c}{0.588}&0.999&\multicolumn{1}{c}{0.736}&0.760&\multicolumn{1}{c}{0.577}&0.794&0.575\\
\midrule
\multicolumn{1}{c}{\multirow{5}{*}{h1 $\xrightarrow{}$m2}}&\multicolumn{1}{c}{96}&\underline{0.214$\pm$0.004}& \multicolumn{1}{c}{\underline{0.299$\pm$0.005}}&\textbf{0.212}& \multicolumn{1}{c}{\textbf{0.298}}&0.218& \multicolumn{1}{c}{0.304}&0.218& \multicolumn{1}{c}{0.301}&0.256& \multicolumn{1}{c}{0.332}&0.219& \multicolumn{1}{c}{0.305}&0.257& \multicolumn{1}{c}{0.345}&0.611& \multicolumn{1}{c}{0.588}&0.239& \multicolumn{1}{c}{0.343}&0.245& \multicolumn{1}{c}{0.322}\\
\multicolumn{1}{r}{}&\multicolumn{1}{c}{192}&\textbf{0.276$\pm$0.006}&\multicolumn{1}{c}{\textbf{0.332$\pm$0.008}}&0.277&\multicolumn{1}{c}{0.338}&0.279&\multicolumn{1}{c}{0.338}&\underline{0.278}&\multicolumn{1}{c}{\underline{0.334}}&0.284&\multicolumn{1}{c}{0.342}&0.280&\multicolumn{1}{c}{0.341}&0.318&\multicolumn{1}{c}{0.380}&0.789&\multicolumn{1}{c}{0.685}&0.320&\multicolumn{1}{c}{0.397}&0.293&0.346\\
\multicolumn{1}{r}{}&\multicolumn{1}{c}{336}&\textbf{0.335$\pm$0.007}&\multicolumn{1}{c}{\textbf{0.367$\pm$0.006}}&\textbf{0.336}&\multicolumn{1}{c}{\underline{0.371}}&0.342&\multicolumn{1}{c}{0.371}&\underline{0.338}&\multicolumn{1}{c}{0.369}&0.357&\multicolumn{1}{c}{0.385}&0.341&\multicolumn{1}{c}{0.376}&0.375&\multicolumn{1}{c}{0.417}&1.469&\multicolumn{1}{c}{0.927}&0.409&\multicolumn{1}{c}{0.453}&0.361&0.382\\
\multicolumn{1}{r}{}&\multicolumn{1}{c}{720}&\textbf{0.429$\pm$0.007}&\multicolumn{1}{c}{\textbf{0.417$\pm$0.005}}&0.435&\multicolumn{1}{c}{0.424}&0.432&\multicolumn{1}{c}{0.419}&\underline{0.431}&\multicolumn{1}{c}{\underline{0.418}}&0.449&\multicolumn{1}{c}{0.433}&0.432&\multicolumn{1}{c}{0.426}&0.480&\multicolumn{1}{c}{0.472}&1.612&\multicolumn{1}{c}{0.957}&0.629&\multicolumn{1}{c}{0.565}&0.460&0.432\\
\multicolumn{1}{r}{}&\multicolumn{1}{c}{Avg.}&\textbf{0.313}&\multicolumn{1}{c}{\textbf{0.353}}&\underline{0.315}&\multicolumn{1}{c}{\underline{0.357}}&0.317&\multicolumn{1}{c}{0.359}&\underline{0.316}&\multicolumn{1}{c}{\underline{0.355}}&0.336&\multicolumn{1}{c}{0.373}&0.318&\multicolumn{1}{c}{0.362}&0.357&\multicolumn{1}{c}{0.403}&1.120&\multicolumn{1}{c}{0.789}&0.399&\multicolumn{1}{c}{0.439}&0.339&0.370\\
\midrule
\multicolumn{1}{c}{\multirow{5}{*}{h2 $\xrightarrow{}$m1}}&\multicolumn{1}{c}{96}&\underline{0.797$\pm$0.006}& \multicolumn{1}{c}{\textbf{0.581$\pm$0.003}}&0.891& \multicolumn{1}{c}{0.587}&0.985& \multicolumn{1}{c}{0.604}&0.933& \multicolumn{1}{c}{0.596}&1.099& \multicolumn{1}{c}{0.654}&{0.815}& \multicolumn{1}{c}{{0.560}}&{\textbf{0.734}}& \multicolumn{1}{c}{\underline{0.578}}&1.032& \multicolumn{1}{c}{0.620}&0.762& \multicolumn{1}{c}{0.567}&1.205& \multicolumn{1}{c}{0.678}\\
\multicolumn{1}{r}{}&\multicolumn{1}{c}{192}&\underline{0.790$\pm$0.004}&\multicolumn{1}{c}{\textbf{0.562$\pm$0.004}}&0.850&\multicolumn{1}{c}{0.583}&0.872&\multicolumn{1}{c}{0.600}&0.889&\multicolumn{1}{c}{0.591}&0.966&\multicolumn{1}{c}{0.625}&0.900&\multicolumn{1}{c}{0.606}&\textbf{0.723}&\multicolumn{1}{c}{\underline{0.594}}&1.176&\multicolumn{1}{c}{0.676}&0.785&\multicolumn{1}{c}{0.588}&1.159&0.670\\
\multicolumn{1}{r}{}&\multicolumn{1}{c}{336}&\textbf{0.748$\pm$0.006}&\multicolumn{1}{c}{\textbf{0.568$\pm$0.005}}&0.853&\multicolumn{1}{c}{0.594}&0.926&\multicolumn{1}{c}{0.614}&{0.827}&\multicolumn{1}{c}{{0.587}}&0.947&\multicolumn{1}{c}{0.629}&0.906&\multicolumn{1}{c}{0.602}&\underline{0.750}&\multicolumn{1}{c}{\underline{0.590}}&1.199&\multicolumn{1}{c}{0.718}&0.767&\multicolumn{1}{c}{0.594}&1.197&0.689\\
\multicolumn{1}{r}{}&\multicolumn{1}{c}{720}&\underline{0.767$\pm$0.002}&\multicolumn{1}{c}{\textbf{0.587$\pm$0.003}}&0.879&\multicolumn{1}{c}{0.616}&0.899&\multicolumn{1}{c}{0.624}&{0.821}&\multicolumn{1}{c}{{0.598}}&0.909&\multicolumn{1}{c}{0.623}&0.866&\multicolumn{1}{c}{0.619}&\underline{0.760}&\multicolumn{1}{c}{\underline{0.592}}&1.373&\multicolumn{1}{c}{0.832}&0.800&\multicolumn{1}{c}{0.627}&1.583&0.784\\
\multicolumn{1}{r}{}&\multicolumn{1}{c}{Avg.}&\underline{0.775}&\multicolumn{1}{c}{\textbf{0.569}}&0.868&\multicolumn{1}{c}{0.595}&0.920&\multicolumn{1}{c}{0.610}&0.867&\multicolumn{1}{c}{0.593}&0.980&\multicolumn{1}{c}{0.632}&0.871&\multicolumn{1}{c}{0.596}&\underline{0.741}&\multicolumn{1}{c}{\underline{0.588}}&1.195&\multicolumn{1}{c}{0.711}&0.778&\multicolumn{1}{c}{0.594}&1.286&0.705\\
\midrule
\multicolumn{1}{c}{\multirow{5}{*}{h2 $\xrightarrow{}$m2}}&\multicolumn{1}{c}{96}&\textbf{0.216$\pm$0.002}& \multicolumn{1}{c}{\textbf{0.302$\pm$0.002}}&0.228& \multicolumn{1}{c}{0.311}&0.235& \multicolumn{1}{c}{0.316}&\underline{0.226}& \multicolumn{1}{c}{\underline{0.310}}&0.238& \multicolumn{1}{c}{0.323}&0.288& \multicolumn{1}{c}{0.366}&0.261& \multicolumn{1}{c}{0.347}&0.821& \multicolumn{1}{c}{0.634}&0.264& \multicolumn{1}{c}{0.496}&0.244& \multicolumn{1}{c}{0.324}\\
\multicolumn{1}{r}{}&\multicolumn{1}{c}{192}&\textbf{0.277$\pm$0.003}&\multicolumn{1}{c}{\textbf{0.336$\pm$0.002}}&0.283&\multicolumn{1}{c}{0.341}&0.287&\multicolumn{1}{c}{0.346}&\underline{0.284}&\multicolumn{1}{c}{\underline{0.343}}&0.295&\multicolumn{1}{c}{0.353}&0.344&\multicolumn{1}{c}{0.375}&0.313&\multicolumn{1}{c}{0.370}&1.732&\multicolumn{1}{c}{1.018}&0.394&\multicolumn{1}{c}{0.452}&0.331&0.374\\
\multicolumn{1}{r}{}&\multicolumn{1}{c}{336}&\textbf{0.335$\pm$0.003}&\multicolumn{1}{c}{\textbf{0.369$\pm$0.003}}&0.343&\multicolumn{1}{c}{0.376}&0.361&\multicolumn{1}{c}{0.391}&\underline{0.338}&\multicolumn{1}{c}{\underline{0.372}}&0.354&\multicolumn{1}{c}{0.385}&0.438&\multicolumn{1}{c}{0.425}&0.401&\multicolumn{1}{c}{0.431}&2.587&\multicolumn{1}{c}{1.393}&0.506&\multicolumn{1}{c}{0.513}&0.386&0.405\\
\multicolumn{1}{r}{}&\multicolumn{1}{c}{720}&\textbf{0.430$\pm$0.001}&\multicolumn{1}{c}{\textbf{0.417$\pm$0.002}}&0.437&\multicolumn{1}{c}{0.424}&0.444&\multicolumn{1}{c}{0.433}&\underline{0.433}&\multicolumn{1}{c}{\underline{0.421}}&0.449&\multicolumn{1}{c}{0.434}&0.611&\multicolumn{1}{c}{0.588}&0.487&\multicolumn{1}{c}{0.472}&3.034&\multicolumn{1}{c}{1.452}&0.822&\multicolumn{1}{c}{0.655}&0.485&0.458\\
\multicolumn{1}{r}{}&\multicolumn{1}{c}{Avg.}&\textbf{0.314}&\multicolumn{1}{c}{\textbf{0.356}}&0.322&\multicolumn{1}{c}{0.363}&0.331&\multicolumn{1}{c}{0.371}&\underline{0.320}&\multicolumn{1}{c}{\underline{0.361}}&0.334&\multicolumn{1}{c}{0.373}&0.420&\multicolumn{1}{c}{0.433}&0.365&\multicolumn{1}{c}{0.405}&2.043&\multicolumn{1}{c}{1.124}&0.496&\multicolumn{1}{c}{0.496}&0.361&0.390\\
\bottomrule
\end{tabular}
\end{table*}
\begin{figure*}[!htbp]
    \centering
    \includegraphics[width=0.7\linewidth]{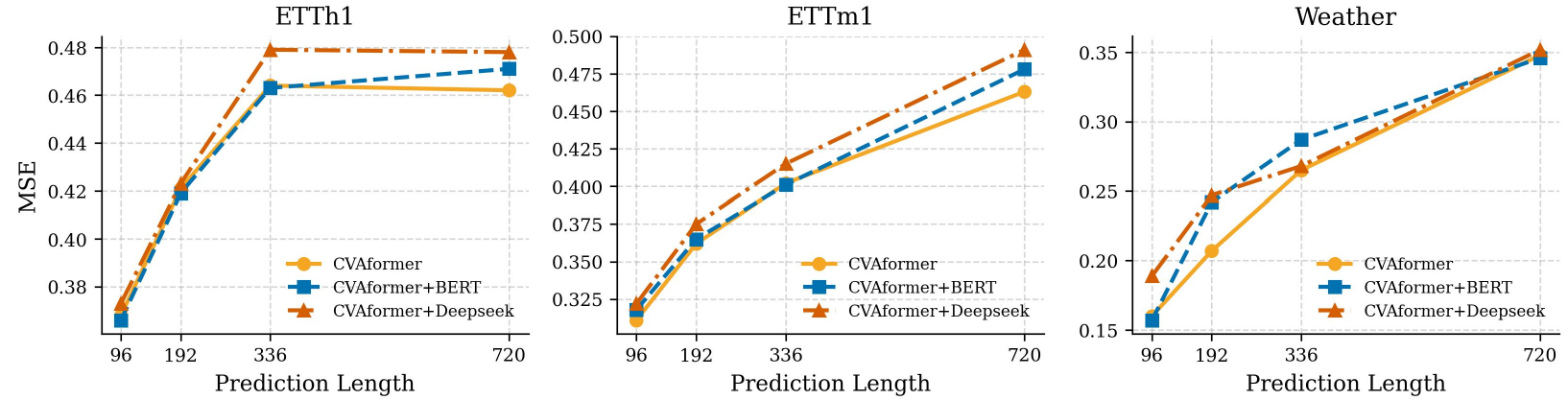}
    \caption{Performance comparison of CVAformer with different LLM backbones: GPT-2, BERT, and DeepSeek-R1-Distill-Qwen-1.5B, on ETTh1, ETTm1, and Weather Datasets.}
    \label{general}
\end{figure*}
\begin{table*}[!htbp]
    \caption{The ablation study for each component of CVAformer in long-term forecasting with ETTh1 and Weather datasets. The best results are highlighted in \textbf{bold}.}
    \label{ablation long}
    \centering
    \tiny
    \begin{tabular}{ccccccccccc|cccccccccc}
    \toprule
         \multicolumn{1}{c}{Dataset}&\multicolumn{10}{c|}{ETTh1}&\multicolumn{10}{c}{Weather}  \\
         \midrule
         \multicolumn{1}{c}{Type}&\multicolumn{2}{c}{CVAformer}&\multicolumn{2}{c}{\textbf{w/o Cau}}&\multicolumn{2}{c}{\textbf{w/o Non}}&\multicolumn{2}{c}{ \textbf{w/o $\mathcal{L}_{\text{invariant}}$}}&\multicolumn{2}{c|}{$\text{CVAformer}_\text{reverse}$}&\multicolumn{2}{c}{CVAformer}&\multicolumn{2}{c}{\textbf{w/o Cau}}&\multicolumn{2}{c}{\textbf{w/o Non}}&\multicolumn{2}{c}{ \textbf{w/o $\mathcal{L}_{\text{invariant}}$}}&\multicolumn{2}{c}{$\text{CVAformer}_\text{reverse}$}\\
         \midrule
         \multicolumn{1}{c}{Metric}&MSE&MAE&MSE&MAE&MSE&MAE&MSE&MAE&MSE&MAE&MSE&MAE&MSE&MAE&MSE&MAE&MSE&MAE&MSE&MAE\\
         \midrule
         \multicolumn{1}{c}{96}&\textbf{0.368}&\textbf{0.388}&0.376&0.399&0.372&0.392&0.375&0.396&0.369&0.388&\textbf{0.160}&\textbf{0.202}&0.165&0.208&0.162&0.204&0.167&0.206&0.160&0.203\\
         \midrule
         \multicolumn{1}{c}{192}&\textbf{0.421}&\textbf{0.421}&0.427&0.426&0.425&0.424&0.427&0.423&0.423&0.420&\textbf{0.207}&\textbf{0.247}&0.217&0.256&0.210&0.248&0.214&0.251&0.206&0.247\\
         \midrule
         \multicolumn{1}{c}{336}&\textbf{0.463}&\textbf{0.444}&0.480&0.452&0.466&0.445&0.467&0.446&0.463&0.443&\textbf{0.265}&\textbf{0.287}&0.272&0.297&0.267&0.286&0.270&0.295&0.264&0.287\\
         \midrule
        \multicolumn{1}{c}{720}&\textbf{0.462}&\textbf{0.462}&0.486&0.473&0.467&0.465&0.474&0.466&0.463&0.464&\textbf{0.348}&\textbf{0.346}&0.354&0.351&0.349&0.347&0.351&0.349&0.349&0.347\\
         \bottomrule
    \end{tabular}
\end{table*}
\begin{figure*}[!htbp]
\vspace{-0.03in}
    \centering
    \includegraphics[width=0.9\linewidth]{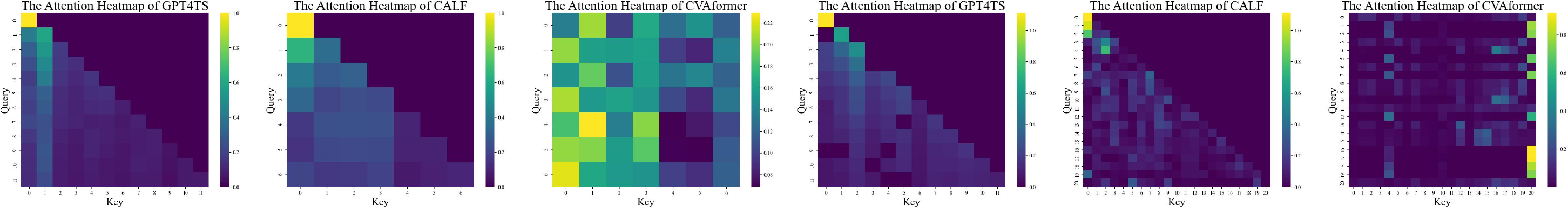}
    \caption{The visualizations of the learned attention maps by GPT4TS \cite{GPT4TS}, CALF \cite{CALF}, and the proposed CVAformer on ETTh1 and Weather datasets, respectively. The first three subfigures illustrate the attention maps learned on the ETTh1 dataset, while the latter three correspond to results on the Weather dataset.}
    \label{ablation}
\end{figure*}
\begin{figure*}[htbp]
    \centering
    \includegraphics[width=0.8\linewidth]{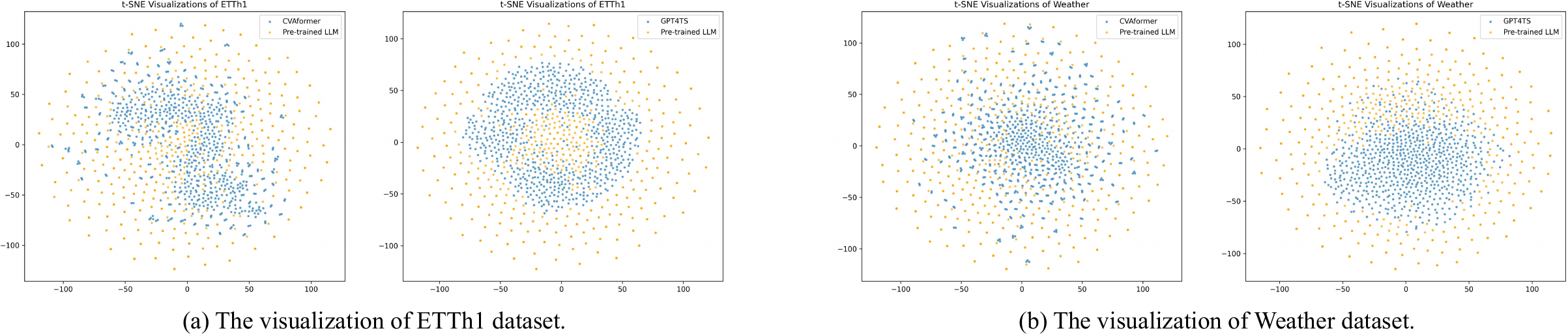}
    \caption{The t-SNE visualization of pre-trained word token embeddings of LLM with temporal tokens of (a) ETTh1 and (b) Weather dataset from GPT4TS \cite{GPT4TS} (right) and our method (left).}
    \label{tsne}
\end{figure*}
\subsection{Model Analysis}
\textbf{Ablation studies on model components.} To evaluate the contributions of individual components within CVAformer, ablation studies are conducted on three key modules. The ablated variants are defined as follows: 1) \textbf{w/o Cau}, which removes the causal disentanglement module, including the decomposition block, the CausalEncoder, and the gating-based intervention. 2) \textbf{w/o Non}, which replaces the non-causal attention module with the standard causal attention used in GPT-style LLMs. 3) \textbf{w/o $\mathcal{L}_{\text{invariant}}$}, which disables the invariant-specific contrastive loss by setting $\lambda_{2} = 0$. It can be found in Table \ref{ablation} that each component contributes significantly to the overall performance of CVAformer. Notably, removing the causal disentanglement leads to an average MSE increase of 3.2\% on ETTh1 and 2.8\% on Weather datasets, underscoring the importance of addressing confounding effects in alignment. Furthermore, removing non-causal attention degrades performance, confirming its effectiveness in variable-level alignment.

To evaluate whether our model depends on variable ordering, we conduct a permutation-sensitivity experiment: randomly permute the variable axis while keeping the model parameters fixed. The results are shown in Table \ref{ablation} with annotation as $\text{CVAformer}_\text{reverse}$. Although NCBlocks are permutation-equivariant by design, the full model is not strictly invariant due to the alignment branch and gating mechanism. This explains why permutation experiments still show observable but reduced variance. The observed changes remain moderate, showing that CVAformer exhibits controlled sensitivity to variable reordering, and the extracted embedding is stable. 

\textbf{Generality.} To evaluate the compatibility of CVAformer with different LLM backbones, representative models such as BERT \cite{bert} and DeepSeek-R1-Distill-Qwen-1.5B \cite{guo2025deepseek} are adopted. These models differ significantly in architecture and training paradigms, allowing us to evaluate whether the proposed framework remains effective across LLM variants. For fair comparisons, non-causal attention modules are removed in all configurations, as BERT \cite{bert} is a bidirectional encoder by design. In our implementation, the word embedding dimension of BERT is 768, while the dimension of DeepSeek-R1-Distill-Qwen-1.5B is 1536. The results, as shown in Fig. \ref{general}, demonstrate that CVAformer maintains competitive performance across different architectures, highlighting its generality. Notably, the BERT-based variant achieves the best results in certain cases, further supporting the effectiveness of non-causal attention in variable-level modeling.

\textbf{The effect of non-causal attention.} Attention maps learned by GPT4TS \cite{GPT4TS}, CALF \cite{CALF}, and the proposed CVAformer on ETTh1 and Weather datasets are visualized in Fig. \ref{ablation}. While GPT4TS \cite{GPT4TS} captures autoregressive dependencies through patch-level causal attention, CALF \cite{CALF} imposes a restrictive temporal ordering during variable-level alignment, contradicting the inherent nature of multivariate series. By comparison, CVAformer utilizes non-causal attention maps to explicitly capture unordered inter-variable dependencies. These patterns better reflect forecasting semantics and demonstrate that non-causal modeling provides superior interpretability and suitability for multivariate relationships.

\textbf{The t-SNE Visualizations of Different Datasets.} To further demonstrate the alignment capability, we visualize the t-SNE distributions comparing the pre-trained word token embeddings of the LLM with the temporal tokens from different datasets produced by GPT4TS \cite{GPT4TS} and CVAformer. Existing LLM-based predictors often overlook the distribution discrepancy between textual and temporal modalities, leading to suboptimal alignment \cite{CALF}. Results clearly show that our approach achieves a more cohesive integration between modalities as it achieves a lower distribution discrepancy. This consistent clustering underscores the robustness of our variable-level alignment and causal disentanglement in bridging the modality gap.

\begin{figure}
    \centering
    \includegraphics[width=0.75\linewidth]{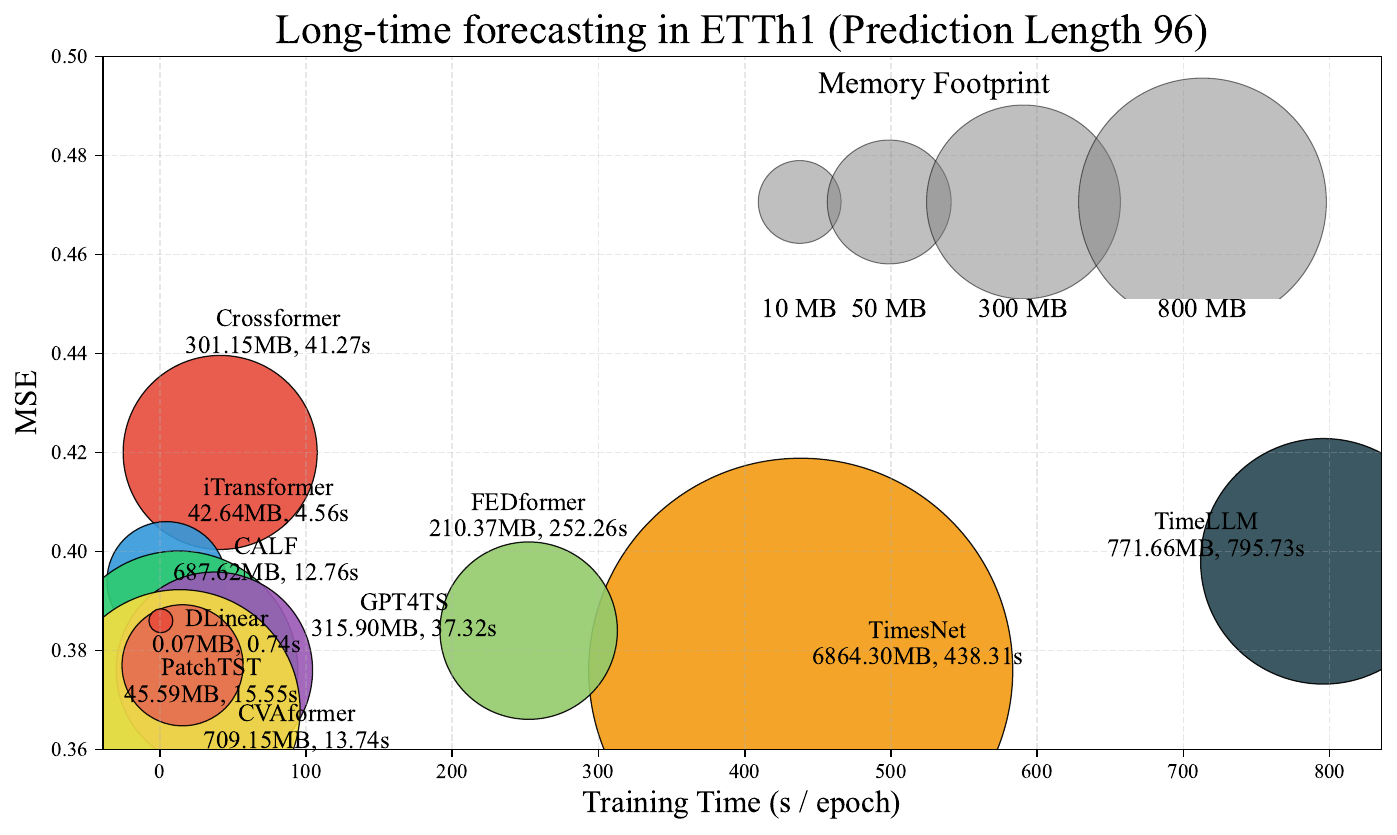}
    \caption{Comparison of model efficiency on ETTh1 ( prediction length = 96). }
    \label{bubble}
    \vspace{-0.1in}
\end{figure}

\textbf{Computational Efficiency Analysis}
To further evaluate the practicality of CVAformer, we conduct a comprehensive comparison of its performance, training efficiency and memory
footprint on the ETTh1 dataset with prediction length as 96. As shown in Fig. \ref{bubble}, CVAformer achieves the lowest MSE of 0.368 while maintaining a moderate training time of 13.74 s/epoch and a compact memory
footprint of 709.15 MB among all baselines. As a large model-based time series forecasting method, CVAformer demonstrates strong performance while maintaining competitive efficiency.
\section{Conclusion and future work}\label{conclusion}
In this paper, a novel framework CVAformer is proposed for time series forecasting that mitigates the modality gap between numerical time series data and the language-based representations used in LLMs via a causally guided variable-level alignment. Unlike patch- or time-step-level methods, CVAformer aligns individual variables with pretrained word embeddings using non-causal attention to capture order-invariant interactions. To mitigate spurious correlations, we introduce a disentanglement mechanism with soft gating that separates dynamic and invariant components. Extensive evaluations across diverse benchmarks demonstrate its superior accuracy and generalization over state-of-the-art LLM-based and traditional baselines. Future research will focus on deeper causal analysis of the alignment process and enhancing model interpretability for real-world applications.
\appendix
\textbf{The contrastive learning for $\mathcal{L}_{\text{invariant}}$.}\label{contrastive}
To enhance the disentanglement between invariant and dynamic components, we employ contrastive learning using a MoCo-style framework \cite{MoCo}. Given an input sample $x_{i}$, two augmentations $a, a' \sim \mathcal{A}$ are applied to generate a query $q_{i} = f(a(x_{i}))$ and a positive key $k_{i} = f(a'(x_{i}))$, where $f$ denotes the projection head. The invariant loss is formulated as:
\begin{equation}
    \mathcal{L}_{\text{invariant}} = \sum_{i=1}^{N}-\text{log}\frac{\text{exp}(q_{i}\cdot k_{i}/\tau)}{\text{exp}(q_{i}\cdot k_{i}/\tau)+\sum_{j=1}^{K_{\text{negative}}}\text{exp}(q_{i}\cdot k_{j}/\tau)},
\end{equation}
where $\tau$ is the temperature hyperparameter and $\{k_{j}\}_{j=1}^{K}$ are negative samples maintained in a dynamic queue. Following \cite{MoCo}, the queue is updated by dequeuing the oldest mini-batch and enqueuing the current representations, enabling a large set of negative samples for robust representation learning.

\textbf{Experimental Details.}\label{Experimental-appendix}
All experiments are implemented in PyTorch and executed on a single NVIDIA A40 GPU. We utilize the Adam optimizer with a learning rate in $\{10^{-3}, 5\times10^{-4}\}$ and a batch size selected from $\{128, 256, 512\}$. The training process is set to 20 epochs. Following \cite{CALF}, we employ the first 6 Transformer layers of a pre-trained GPT-2 as the backbone. Low-Rank Adaptation (LoRA) is applied for fine-tuning, with the rank $r \in \{4, 8\}$ and $\alpha \in \{16, 20\}$. The PCA dimension is set to 500, and dropout rates are chosen from $\{0.01, 0.3, 0.5\}$. For series decomposition, the kernel size is selected from $\{1, 3, \dots, 49\}$ based on dataset characteristics. Non-causal attention is selectively activated for high-dimensional datasets to better capture variable-level dependencies. The total loss is a weighted combination of $\mathcal{L}_{\text{sup}}$, $\mathcal{L}_{\text{consistency}}$, and $\mathcal{L}_{\text{invariant}}$. The hyperparameters are tuned with $\lambda_{1} \in \{0.8, 1\}$ and $\lambda_{2} \in \{0.01, 0.02, 0.05\}$. Our implementation is built upon the TimesNet \cite{TimesNet} benchmark to ensure fair comparison under standard configurations.

\refstepcounter{section}
\bibliographystyle{IEEEtran}
\bibliography{main_new}


\end{document}